\documentclass[final,5p,times,twocolumn]{elsarticle}
\usepackage[utf8]{inputenc} 
\usepackage[T1]{fontenc}    
\usepackage{url}            
\usepackage{booktabs}       
\usepackage{amsfonts}       
\usepackage{nicefrac}       
\usepackage{microtype}      
\usepackage{color}
\usepackage{graphicx}
\usepackage{float}
\usepackage{multirow}
\usepackage{amsmath}
\usepackage{verbatim}
\usepackage{amsmath}
\usepackage{rotating} 
\usepackage{lineno}
\usepackage{amsmath}
\usepackage{amsfonts}
\usepackage{amssymb}
\usepackage{algorithm}
\usepackage{algorithmic}
\usepackage{array}
\usepackage{bm}
\usepackage{graphicx}
\usepackage{cleveref}
\usepackage{footmisc}
\usepackage{color}
\usepackage{subcaption}
\usepackage{bbm}
\usepackage{bbding}
\usepackage{tablefootnote}
\usepackage{diagbox}
\usepackage{xcolor}
\setcitestyle{round,authoryear}

\journal{ISPRS Journal of Photogrammetry and Remote Sensing}

\newcolumntype{C}[1]{>{\centering\arraybackslash}p{#1}}

\def\eg{\emph{e.g}.}

\def\etal{\emph{et al}.}
\def\ie{\emph{i.e}.}

\begin{document}

\begin{frontmatter}

\title{A Large-scale Climate-Aware Satellite Image Dataset
\\
for Domain Adaptive Land-Cover Semantic Segmentation
}

\author[1,2]{Songlin Liu \corref{*}}
\author[3,4]{Linwei Chen \corref{*}}
\author[2]{Li Zhang}
\author[5]{Jun Hu}
\author[3,4]{Ying Fu \corref{**}}
\address[1]{School of Remote Sensing Information Engineering, Wuhan University, 430079, Wuhan, Hubei, China}
\address[2]{State Key Laboratory of Geo-information Engineering, 710054, Xi’an, Shaanxi, China}
\address[3]{Advanced Research Institute of Multidisciplinary Science, Beijing Institute of Technology, 100081, Beijing, China}
\address[4]{School of Computer Science and Technology, Beijing Institute of Technology, 100081, Beijing, China}
\address[5]{School of Electronics and Communication Engineering, Sun Yat-sen University, 510006, Guangzhou, Guangdong, China}

\begin{abstract}
{
A few well-annotated datasets for land-cover semantic segmentation have recently been introduced to advance the field of earth observation technologies. 
However, these datasets overlook the significant diversity among geographic areas with different climates, which can greatly impact and diversify land cover. 
Consequently, this leads to a domain gap in remote sensing images and severe performance degradation of the segmentation models.
To enhance land-cover semantic segmentation with improved generalization ability, we conducted the first investigation into the impact of climate on this task. 
In this paper, we present a unique large-scale Climate-Aware Satellite Images Dataset (CASID) specifically designed for domain adaptive land-cover semantic segmentation. 
It consists of 980 satellite images with a size of 5,000 $\times$ 5,000 pixels, 
collected from 30 different regions around Asia, covering over 24,500 square kilometers. 
These images are gathered from four distinct climate zones, namely temperate monsoon, subtropical monsoon, tropical monsoon, and tropical rainforest.
It includes four sub-datasets/domains, each representing one of the aforementioned climate zones. 
This characteristic makes CASID the first climate-aware land-cover semantic segmentation dataset with multiple domains. 
Additionally, we provide a comprehensive analysis of the samples from the four climate zones, emphasizing differences in global image features, image texture, category distribution, spectral value, and object shape. 
These analyses offer valuable insights for subsequent research in this field. 
Moreover, we conduct extensive experiments to evaluate the latest semantic segmentation and unsupervised domain adaptation methods on the CASID dataset. 
These results serve as a robust baseline for future research endeavors. 
Our dataset will be made publicly available soon at the following link: \url{https://github.com/Linwei-Chen/CASID}.

}
\end{abstract}


\end{frontmatter}

\section{Introduction}
Land-cover semantic segmentation, which aims to classify the land-cover type for each pixel in remote sensing images, { plays a crucial role in enhancing our understanding of geographical and ecological environments~\citep{factseg, 2020gid}. 
It finds widespread applications in various remote sensing domains}, such as urban planning~\citep{2021urban}, precision agriculture~\citep{2019agricultural}, food insecurity analysis~\citep{2021impact}, and environmental protection~\citep{2021wetland}.

Existing mainstream semantic segmentation methods for remote sensing images~\citep{deeplabv3plus,pspnet, fcn2015, factseg, 2020aerialDCNN, 2022unetformer, 2021attentionfused} are predominantly based on deep learning, which is a data-driven technology. 
These methods heavily rely on the quality and quantity of the dataset to achieve high accuracy. Consequently, datasets play a crucial role in the advancement of land-cover semantic segmentation for earth observation.
To foster progress in this field, several remote sensing datasets with well-annotated pixel-wise labels for land-cover semantic segmentation have been recently introduced, such as the Zurich Summer dataset~\citep{2015Zurich}, DeepGlobe dataset~\citep{2018deepglobe}, and Gaofen Image Dataset (GID)~\citep{2020gid}.

However, these datasets are primarily designed for standard land-cover semantic segmentation and fail to consider the significant diversity among different geographic areas. 
This oversight leads to a domain gap and can cause severe degradation of deep learning models~\citep{2021loveda}. 
{
In particular, land cover is profoundly influenced and diversified by climate factors~\citep{dale1997relationship, reidmiller2019fourth}. 
The climate serves as a major determinant of vegetation distribution~\citep{prentice1990bioclimatic, woodward2004global, salazar2007climate}. 
For instance, areas within the tropical rainforest climate zone are typically covered by dense rainforests, whereas the temperate monsoon climate zone often exhibits bare ground surfaces with fewer trees, owing to differences in temperature and precipitation.
}
Therefore, the distribution and pixel spectra of the forest category in the temperate monsoon climate zone can differ significantly from those in the tropical rainforest climate zone.

{
Moreover, both climate and vegetation have a significant impact on water partitioning, consequently influencing rivers and bodies of water~\citep{2016quantifying, 2022climate}. Additionally, climate plays a crucial role in shaping the design of buildings, constructions, and roads, as they are influenced by temperature and precipitation considerations~\citep{golany1996urban, givoni1998climate}. 
As a result, there is a substantial disparity in land-cover characteristics among different climate zones, leading to the climate domain gap.
}

Due to the climate domain gap, optimal land-cover semantic segmentation models learned from a certain climate zone may quickly lose their effectiveness on new images captured from different climate zones. 
This problem significantly hampers the practical application of land-cover semantic segmentation. 
However, there is a lack of relevant research investigating the impact of climate on land-cover semantic segmentation. 
To enhance the model's generalizability for land-cover semantic segmentation, an appropriate climate domain adaptive dataset is urgently needed.

In this paper, we explore the impact of climate on land-cover semantic segmentation for the first time. 
{
To accomplish this, we present a distinctive large-scale Climate-Aware Satellite Images Dataset (CASID) designed specifically for domain adaptive land-cover semantic segmentation. 
CASID comprises image samples with large image sizes from four representative climate zones: temperate monsoon (TemMs), subtropical monsoon (SubMs), tropical monsoon (TroMs), and tropical rainforest (TroRf), abbreviated for brevity.
These climate zones form four sub-datasets/domains based on their climate types, making CASID the first land-cover semantic segmentation dataset to incorporate multiple climate domains.
}

\begin{figure}[t!]
\begin{center}
\scalebox{.66}{
\begin{tabular}{c|ccc}
\hspace{-3mm}
\includegraphics[width=0.44\linewidth]{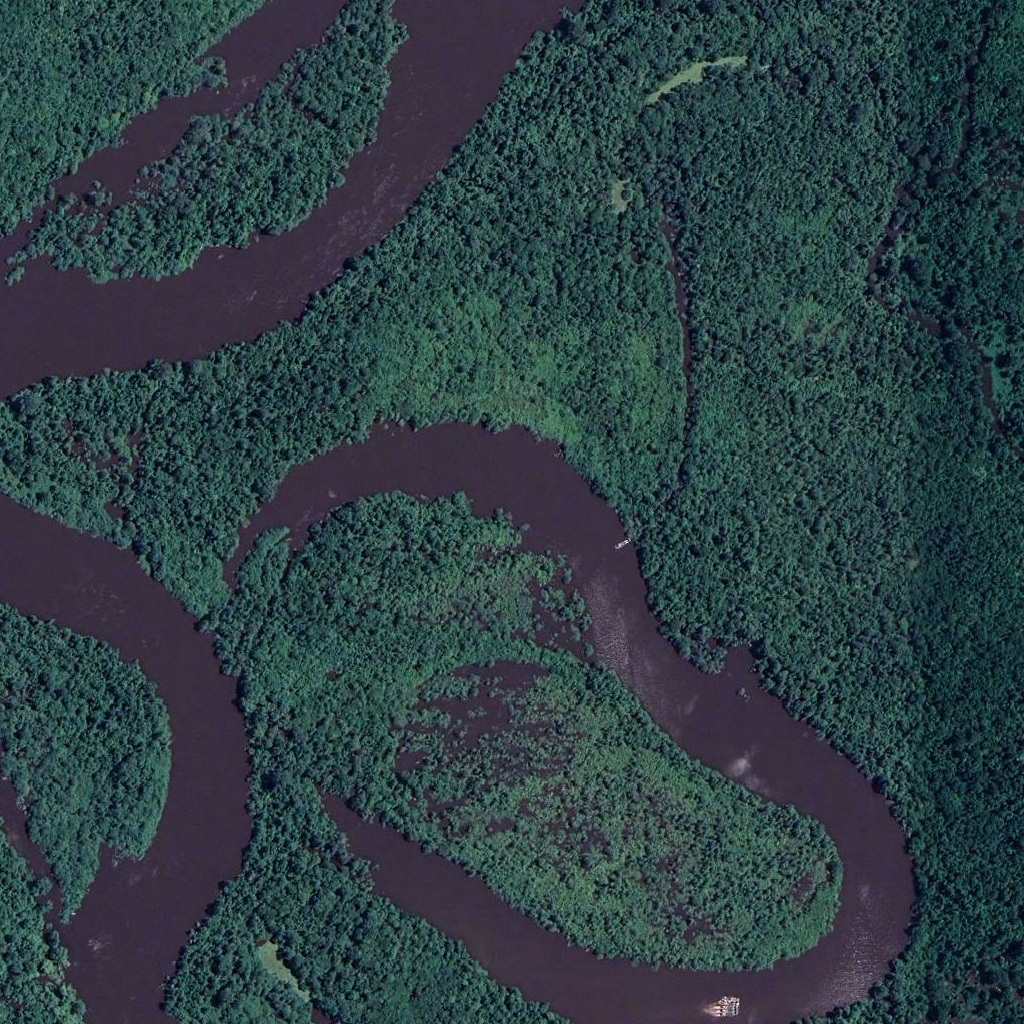}
&\includegraphics[width=0.44\linewidth]{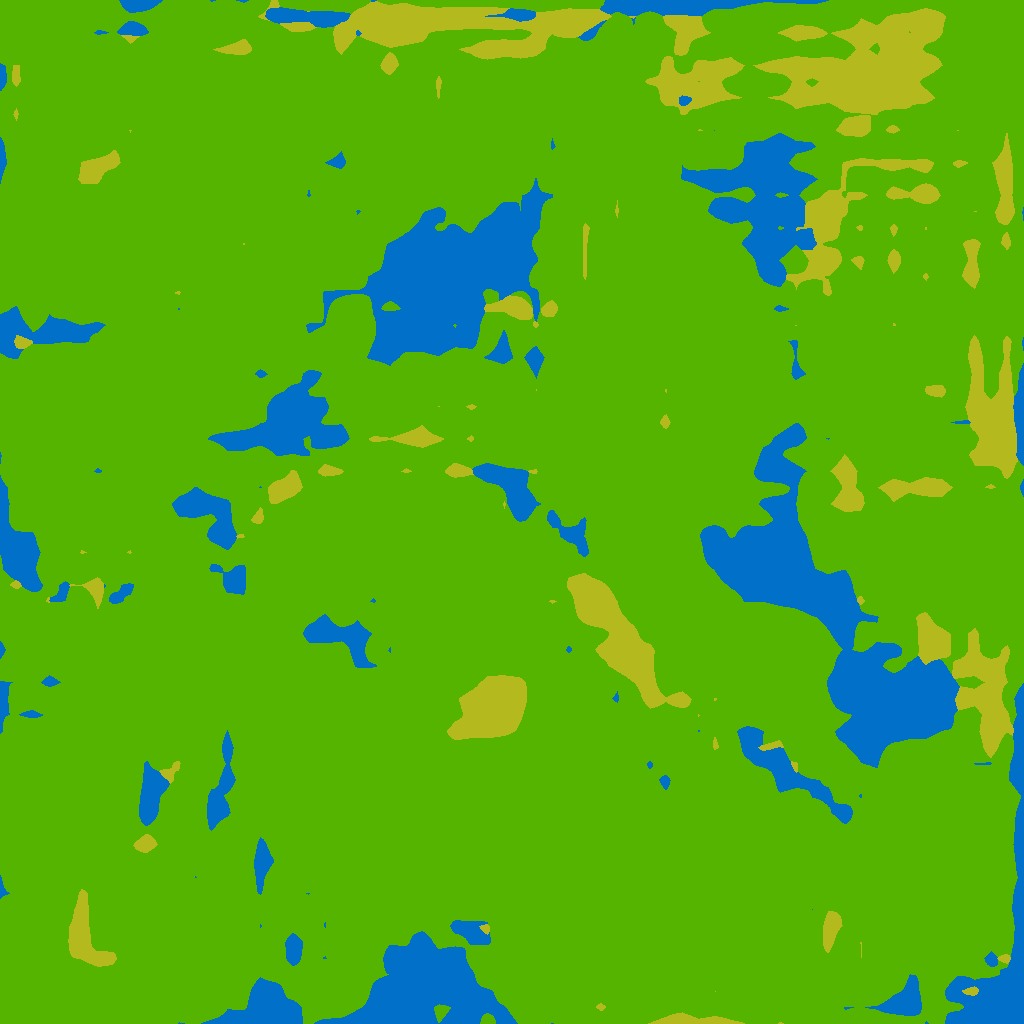}\hspace{-3mm}
&\includegraphics[width=0.44\linewidth]{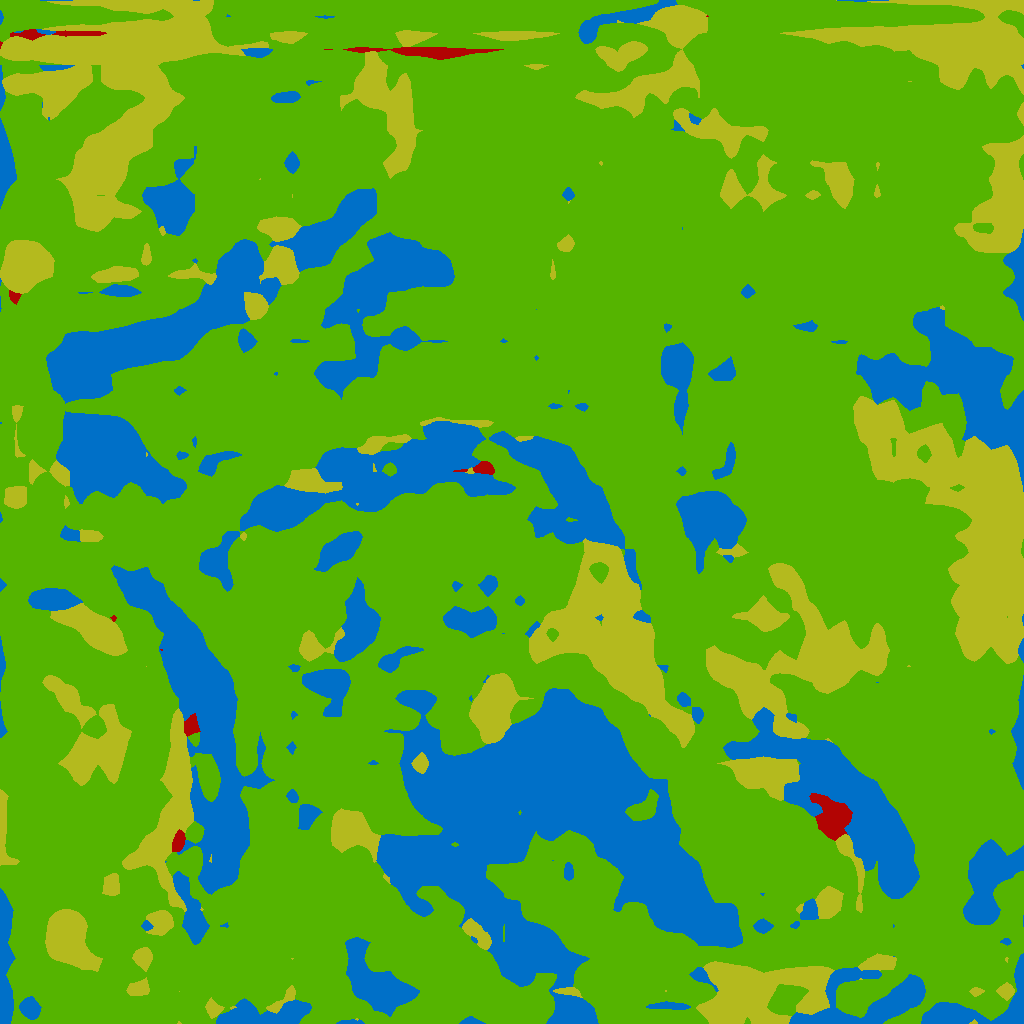}\hspace{-3mm}
\\
(a) Image from TroRf & (b) TemMs $\rightarrow$ TroRf & (c) SubMs $\rightarrow$ TroRf
\\
\hspace{-3mm}
\includegraphics[width=0.44\linewidth]{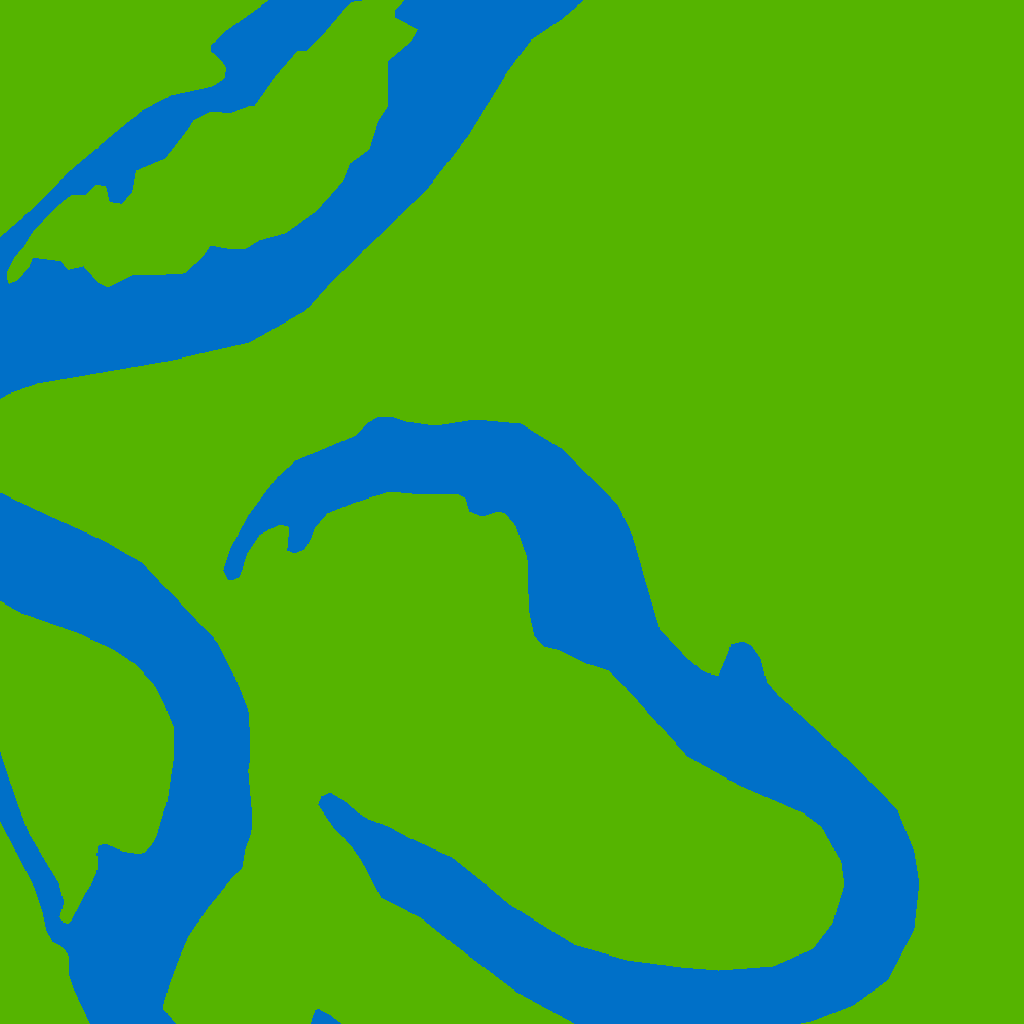}
&\includegraphics[width=0.44\linewidth]{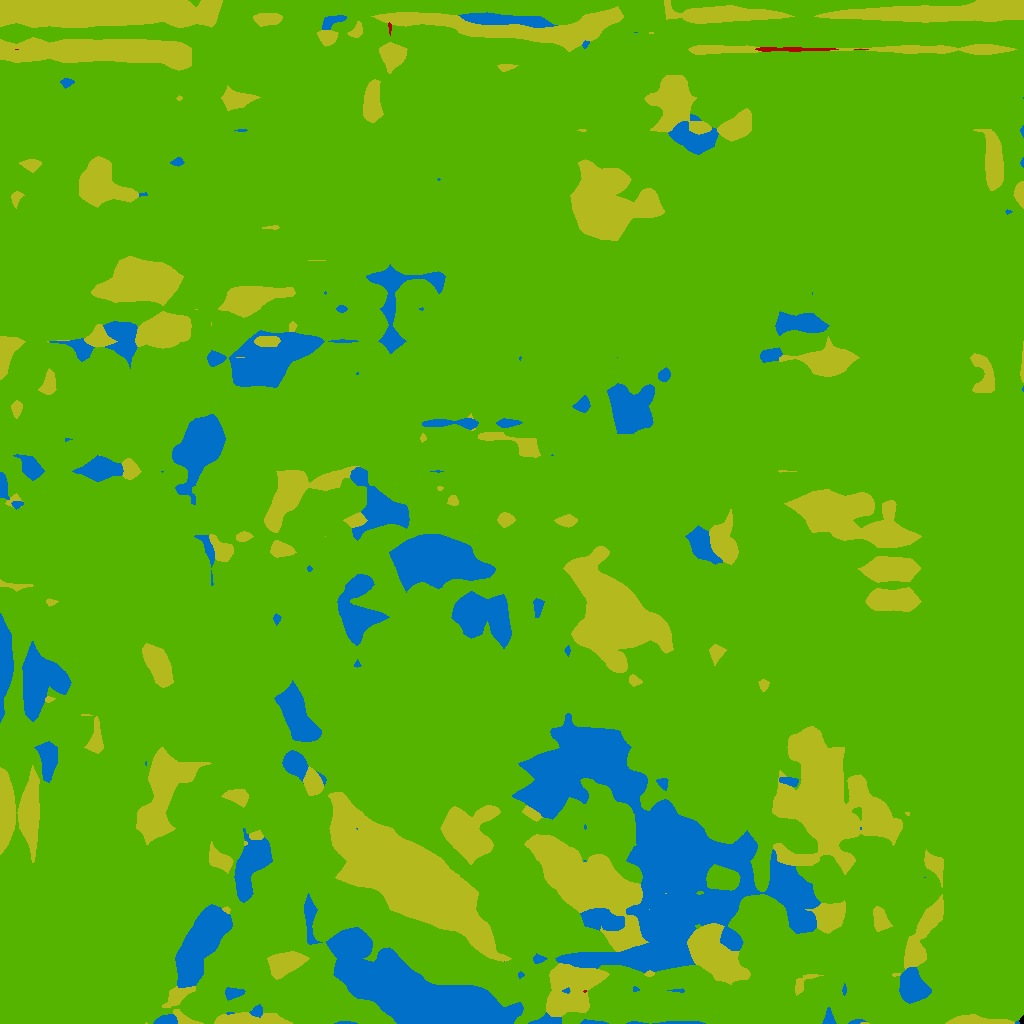}\hspace{-3mm}
&\includegraphics[width=0.44\linewidth]{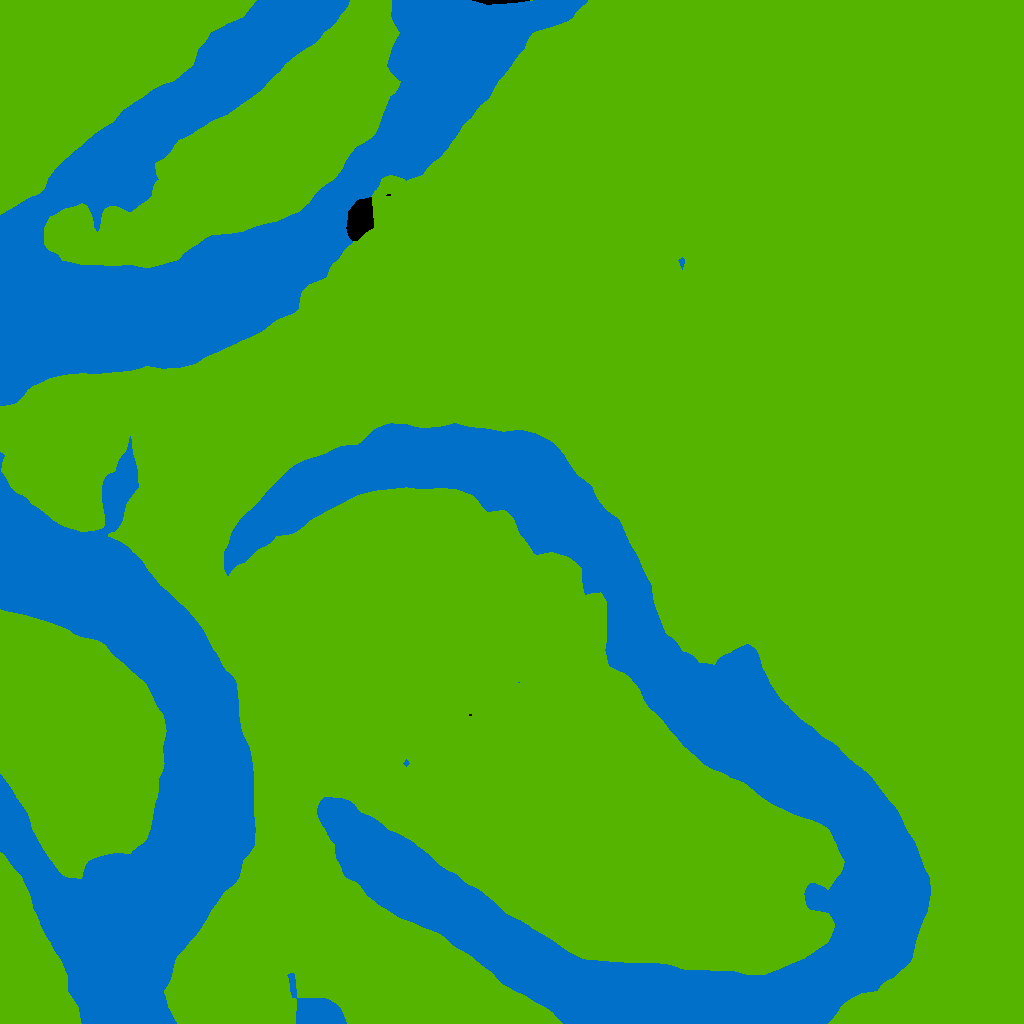}\hspace{-3mm}
\\
(d) Ground Truth & (e) TroMs $\rightarrow$ TroRf & (f) TroRf $\rightarrow$ TroRf 
\\
\end{tabular}
}
\caption{
Visualized results of semantic segmentation model~\citep{deeplabv2}.
The ``A $\rightarrow$ B" notation indicates that the model is trained on domain A and evaluated on domain B.
Due to the climate domain gap, models trained on TemMs, SubMs, and TroMs fail to predict accurate results on TroRf climate zone.
}
\label{fig:intro}
\end{center}
\end{figure}

{
To enhance diversity and increase the challenge of our CASID, we gathered 980 satellite images at a size of $5,000\times 5,000$ from 30 different regions around Asia. 
These regions encompass densely populated urban areas, sparsely populated suburbs, and rural zones, covering a total area of over 24,500 square kilometers. 
For each satellite image, we provide accurate pixel-wise semantic segmentation labels for five categories: background, building, forest, road, and water.
The dataset production process ensures the high diversity and complexity of CASID. Consequently, as shown in Table~\ref{tab:crossdomain}, the widely-used semantic segmentation method for domain adaptation, DeepLabV2~\citep{deeplabv2}, achieves only 39.9$\sim$60.9 mIoU on different subsets. 
When evaluated across climate zones on CASID, the performance degrades catastrophically. 
For instance, the model trained on the temperate monsoon climate zone achieves only 12.2 mIoU on the tropical rainforest climate zone. The cross-domain results depicted in Figures~\ref{fig:intro}(b), (c), and (e) visually confirm these quantitative findings.
These results demonstrate that CASID can serves as a challenging benchmark for validating and developing robust methods in land-cover semantic segmentation and unsupervised domain adaptation.
}

\begin{table}[t!]
\color{black}
\vspace{-0mm}
\caption{
\color{black}
Results of semantic segmentation model~\citep{deeplabv2} trained and evaluated on images from different climate zone.
The model learned from a certain climate zone exhibits severe performance degradation when evaluated on images from different climate zone.
}
\vspace{-0mm}
\begin{center}
\scalebox{0.8}{
\begin{tabular}{l|c|c|c|cc}
\toprule[1.5pt]
\diagbox{Train on}{mIoU}{Test on} & TemMs set & SubMs set&TroMs set& TroRf set \\
\midrule
TemMs set & \bf 39.9 &9.4 & 10.9 & 12.2\\ 
SubMs set & 10.0 &\bf 60.9 & 12.8 & 15.3 \\
TroMs set & 12.2 & 12.8 &\bf 56.5 & 9.1 \\
TroRf set & 8.5 & 10.2 & 12.8 &\bf 49.0\\
\bottomrule[1.5pt]
\end{tabular}
}
\end{center}
\label{tab:crossdomain}
\end{table}

Besides, to shed light on the domain gap caused by climate, { we conducted an analysis of images from these four typical climate zones. 
The analysis reveals large differences in global image features, image texture, category distribution, spectral value, and object shape between different climate zones. 
These differences contribute to the domain gap and the degradation of optimal models trained on certain climate zones.
Furthermore, we conduct experiments with a series of state-of-the-art semantic segmentation methods~\citep{fcn2015, pspnet, deeplabv3plus,unet,unet++, factseg, hrnet, factseg, 2020foreground, linknet, 2021segformer, poolformer, segnext} on the CASID dataset.
And we also exploit the latest unsupervised domain adaptation methods~\citep{adaptseg, CBST, CLAN, PyCDA, FADA, 2020ASA, 2022daformer} to minimize the performance degradation caused by the climatic domain gap.
These results are informative and can serve as a solid baseline for the development of land-cover semantic segmentation.
}

{\color{black}
To summarize, our contributions are as follows:
\begin{itemize}
\item { We have conducted the first investigation into the impact of climate on land-cover semantic segmentation and introduced CASID, a unique climate-aware satellite image dataset with multiple domains.} CASID not only benefits the research community in advancing algorithms with higher generalization ability but also has the potential to become a challenging benchmark for designing robust vision algorithms beyond the field of remote sensing.

\item We analyze samples collected from four typical climate zones and statistically show their differences in global image feature, image texture, category distribution, spectral value, and object shape. These analyses are valuable for inspiring subsequent research.
\item We evaluate the latest semantic segmentation methods and unsupervised domain adaptation methods on the CASID dataset, which can serve as a solid baseline for future research.

\end{itemize} 
} 

The remainder of the paper is organized as follows:
In Section~\ref{sec:related_works}, we introduce the related works.
{
In Section~\ref{sec:dataset}, we provide a detailed description of CASID, including its properties, along with a series of statistical analyses.
In Sections~\ref{sec:ss} and~\ref{sec:uda}, we present the experimental details and results for land-cover semantic segmentation and unsupervised domain adaptation, respectively.
}
Finally, we conclude our work in Section~\ref{sec:conclusion}.

\begin{table*}[t!]
\color{black}
\vspace{-0mm}
\caption{
\color{black}
Comparison between our CASID and the main land-cover datasets for semantic segmentation (SS) and unsupervised domain adaptation (UDA). 
}
\begin{center}
\scalebox{0.728}{
\begin{tabular}{lccccrccccccc}
\toprule[1.5pt]
Dataset & Year &Resolution (m)  &Area (km$^2$)&  Image size & Images & Categories & Domains & SS Task & UDA Task\\
\midrule
ISPRS Vaihingen & 2013 &0.09  & 1.4 & $1887\sim 3816$ &33 & 6 & 1 & $\checkmark$ \\ 
ISPRS Postdam & 2013 &0.05 & 3.4 & $6000\times 6000$ &38 & 6 & 1 & $\checkmark$\\
Zurich Summer \citep{2015Zurich}& 2015 &0.6 & 9.4 & $1000 \times 1150 $ &20 & 8 & 1 & $\checkmark$\\

USGS SIRI-WHU \citep{2015usgs}& 2015 & 0.6 & 33.50 & $9000 \times 10000$ &1 & 4 & 1 & $\checkmark$\\

EvLab-SS \citep{2017evlab} & 2017 & $0.1\sim 2 $ & 722.0 & $4500\times 4500 $ &60 & 11 & 1 & $\checkmark$ \\
SpaceNet~\citep{2018spacenet} & 2017 & 0.5 & 2544.0 & $406\sim 439$ & 6000 & 2 & 1 & $\checkmark$\\
DeepGlobe \citep{2018deepglobe} & 2018 & 0.5 & 1716.9& $2448 \times 2448 $ &1146 & 7 & 1 & $\checkmark$\\
Zeebruges~\citep{2018Zeebruges} & 2018 & 0.05 & 1.8 & $10000 \times 10000$ & 7 & 8 & 1 & $\checkmark$\\ 
AIRS~\citep{2019AIRS} & 2019 & 0.07 & 475.0 & $10000 \times 10000$ & 1047 & 2 & 1 & $\checkmark$\\ 
GID \citep{2020gid} &2020 &4 &75900.0&$6800\times 7200$ &150 & 5 & 1 & $\checkmark$\\ 
LandCover.ai\citep{2021landcover} & 2021 & $0.25\sim 0.5$ & 216.3 & $4200\sim 9500$ & 41 & 3 & 1 & $\checkmark$\\ 
LoveDA \citep{2021loveda} &2021 &0.3 &536.2& $1024\times 1024$ &5987 & 7 & 2 & $\checkmark$ & $\checkmark$ \\ 
OpenEarthMap~\citep{2023openearthmap} & 2023 &0.25$\sim$0.5 & 799.0& $1024\times 1024$ &5000 &8 &6 &$\checkmark$ &$\checkmark$ \\
\midrule
CASID (Ours) & 2023 & 1 & 24500.0 & $5000 \times 5000$ &980  & 5 & 4 & $\checkmark$ & $\checkmark$\\
\bottomrule[1.5pt]
\end{tabular}}
\end{center}
\label{tab:datasets}
\end{table*}

\begin{figure*}[t!]
\begin{center}
\scalebox{1.0}{
\begin{tabular}{cccc}
\includegraphics[width=0.98\linewidth]{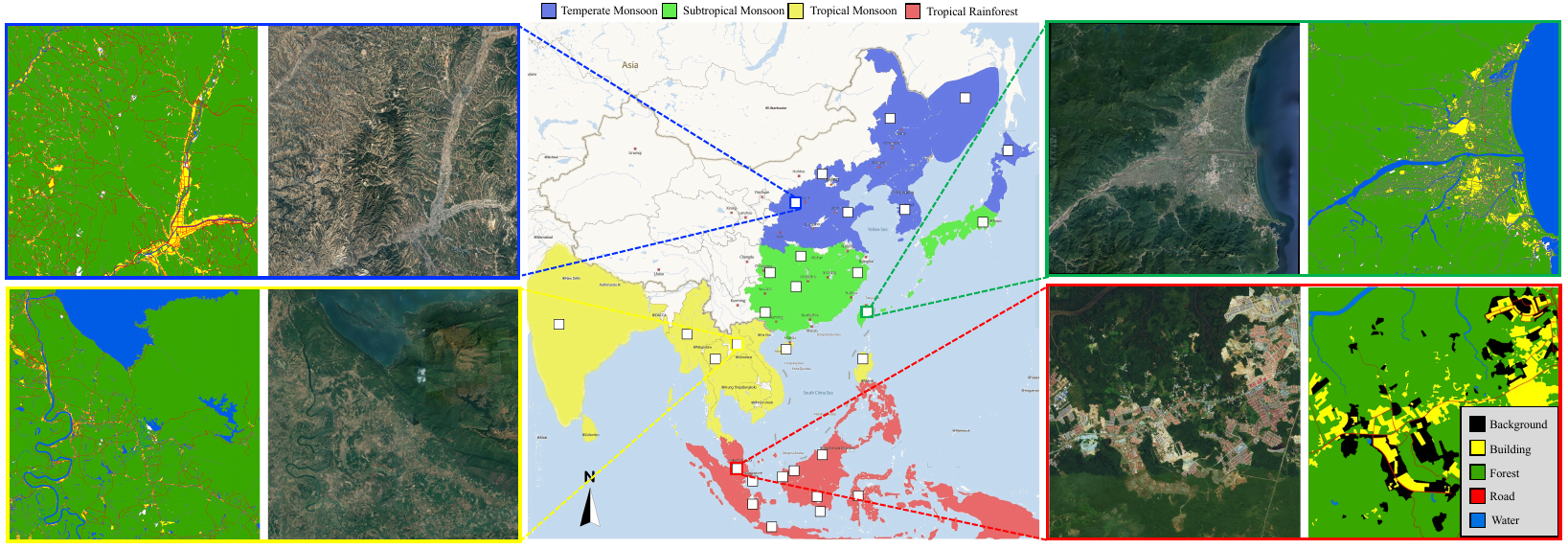} \\
\end{tabular}
}
\caption{
The distribution of geographical locations of images in CASID. 
The ``$\Box$” indicates the sampling areas.
The blue, green, yellow, and red in the climate map indicate four typical climates, \ie, temperate monsoon, subtropical monsoon, tropical monsoon, and tropical rainforest~\citep{1980worldclimate}.
We provide visual examples of representative images from the four climate zones along with their corresponding pixel-wise land-cover labels.
}
\label{fig:CASID}
\end{center}
\end{figure*}

\begin{table*}[t!]
\begin{center}
\vspace{-0mm}
\caption{
The division of the CASID dataset.
Images are collected from 30 regions, covering 4 typical climate zones.
For each climate zone, we choose two regions as the test set and leave the rest as the train set.
The train set and test set account for around 70\% and 30\% of the whole dataset, respectively. 
}
\label{tab:cities}
\vspace{-0mm}
\scalebox{0.88}{
\begin{tabular}{c|c|c|c|c}
\toprule[1.5pt]
Climate & Region & Images & Train/Test split& Subtotal\\
\midrule
\multirow{6}{*}{Temperate monsoon}
& Hokkaido, Japan & 21 
& \multirow{5}{*}{Train: 201} & \multirow{7}{*}{284} \\

\multirow{6}{*}{(TemMs set)}
& Seoul, South Korea& 55 & \\
& Zhangjiakou, Hebei, China& 35 & \\
& Weifang, Shandong, China& 48& \\
& Daqing, Heilongjiang, China& 42& \\

\cmidrule{2-4}
& Belkhabarovsk, Russia& 35 & \multirow{2}{*}{Test: 83}\\
& Lyuliang, Shanxi, China& 48 & \\

\midrule 
\multirow{6}{*}{Subtropical monsoon}
& Yilan, Taiwan, China & 35
& \multirow{5}{*}{Train: 196} & \multirow{7}{*}{284}\\

\multirow{6}{*}{(SubMs set)}
& Jingmen, Hubei, China& 39 &\\
& Baise, Guangxi, China& 38 &\\
& Chongqing, China& 48 &\\
& Jinhua, Zhejiang, China &36 &\\

\cmidrule{2-4}
& Tokyo, Japan& 56 & \multirow{2}{*}{Test: 88}\\
& Shaoyang, Hunan, China& 32 &\\

\midrule
\multirow{5}{*}{Tropical monsoon}
& Wanning, Hainan, China& 22
& \multirow{4}{*}{Train: 136} & \multirow{6}{*}{202}\\
\multirow{5}{*}{(TroMs set)}
& Chiang Mai, Thailand& 36 & \\
& Vientiane, Laos& 36 & \\
& Bhopal, India&42 & \\

\cmidrule{2-4}
& Around Laoag, Philippines& 30 & \multirow{2}{*}{Test: 66}\\
& NayPyitaw, Myanmar& 36 & \\

\midrule
\multirow{9}{*}{Tropical rainforest}
& Kuala Lumpur, Malaysia& 24
& \multirow{8}{*}{Train: 148} & \multirow{10}{*}{210}\\

\multirow{9}{*}{(TroRf set)}
& Lorilindu National Park, Indonesia& 19 & \\
& Linkcuo Port, Singapore& 14 & \\
& Kuching, Malaysia& 36 & \\
& Mandor, Indonesia& 24 & \\
& Guaralupa, Indonesia& 13 & \\
& Klaa, Indonesia& 8 & \\
& Palembang, Indonesia& 10 & \\

\cmidrule{2-4}
& Pharaangkaraya, Indonesia& 26 & \multirow{2}{*}{Test: 62} \\
& Seri Begawan, Malaysia& 36 & \\

\midrule
\multirow{2}{*}{Total} &Train: 22 regions &\multirow{2}{*}{30 regions} & Train: 681 images & \multirow{2}{*}{980 images} \\
&Test: 8 regions & & Test: 299 images & \\

\bottomrule[1.5pt]
\end{tabular}}
\end{center}
\end{table*}

\begin{figure*}[t!]
\begin{center}
\scalebox{1.0}{
\begin{tabular}{cccc}
\includegraphics[width=0.95\linewidth]{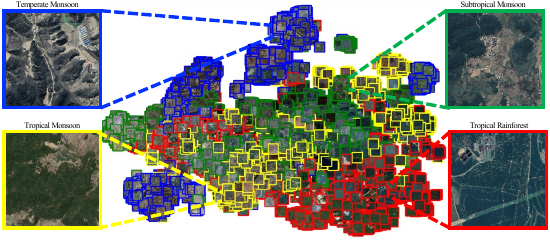} \\
\end{tabular}
}

\caption{
The visualization of global image features analysis on CASID.
We employ two-dimensional t-SNE~\citep{2008tsne} for visualization. 
Colored squares indicate global image features of $1024\times 1024$ image patches sampled from four climate domains in CASID.
To extract the global image features from the image patches, we utilize InceptionV3~\citep{InceptionV3}.
The colors blue, green, yellow, and red represent image patches from the following climate domains: temperate monsoon, subtropical monsoon, tropical monsoon, and tropical rainforest, respectively.
{
After applying t-SNE, we observe that the distribution of image features from the four climate domains closely resembles the distribution of geographical locations shown in Figure~\ref{fig:CASID}.
}
This demonstrates the correlation between land cover and climate zone.
}
\label{fig:dataset_tSNE}
\end{center}
\end{figure*}

\begin{figure}[t!]
\color{black}
\begin{center}
\scalebox{1.0}{
\begin{tabular}{cccc}
\includegraphics[width=0.95\linewidth]{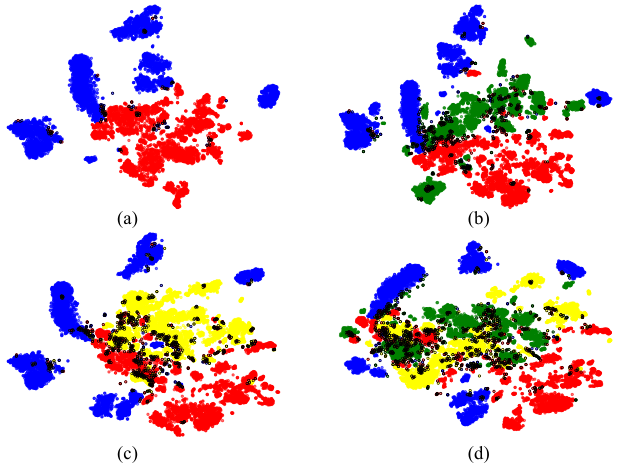} \\
\end{tabular}
}
\caption{
The visualization of global image features analysis on CASID.
We use two-dimensional t-SNE~\citep{2008tsne} for visualization. 
The symbol ``$\circ$" indicates the marginal data, which are predicted by the k-Nearest Neighbors (k-NN) algorithm.
If a data point's nearest neighbor belongs to a different category, then that data point can be considered one of the marginal points, \ie, they are at the boundary of different domains.
Zoom for better details.
}
\label{fig:climate_tSNE}
\end{center}
\end{figure}

\begin{figure}[h!]
\color{black}
\begin{center}

\scalebox{0.828}{
\begin{tabular}{cccc}
\hspace{-2.918mm}
\includegraphics[width=0.2918\linewidth]{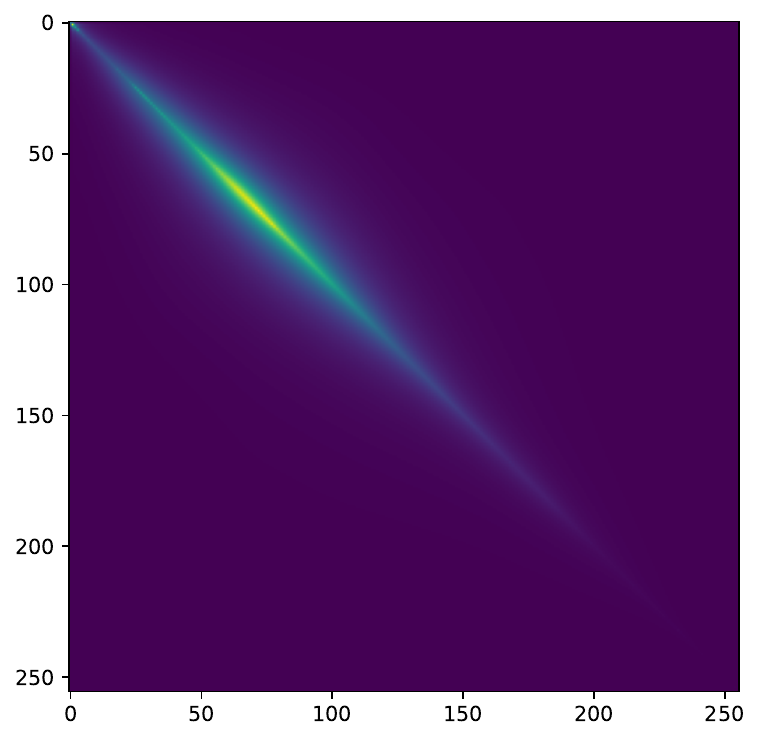}\hspace{-2.918mm}
&\includegraphics[width=0.2918\linewidth]{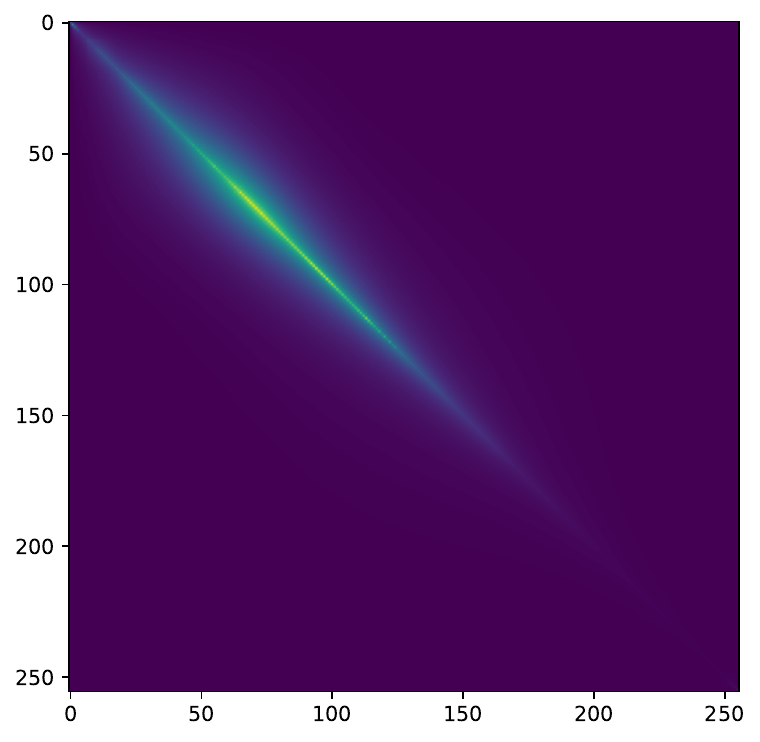}\hspace{-2.918mm}
&\includegraphics[width=0.2918\linewidth]{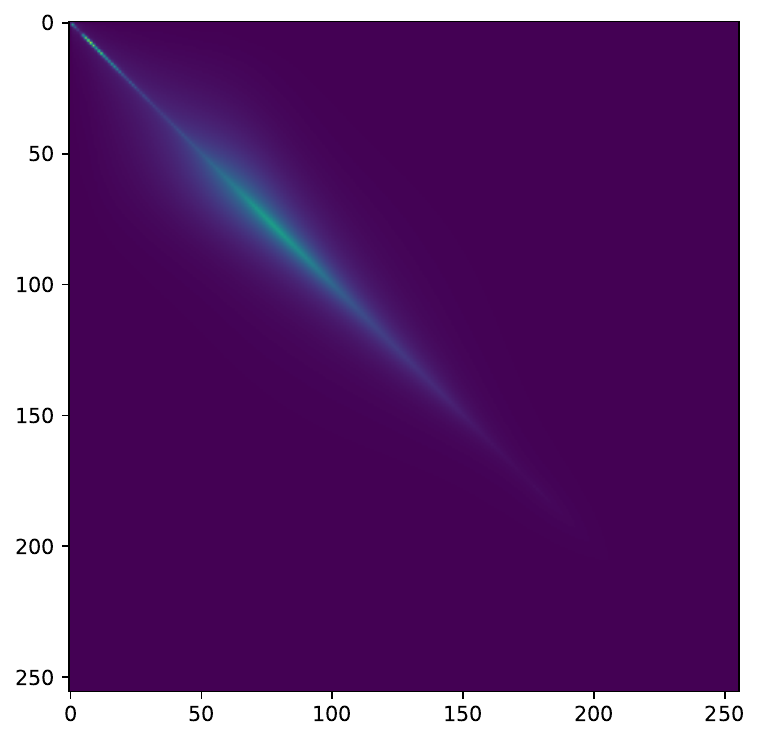}\hspace{-2.918mm}
&\includegraphics[width=0.2918\linewidth]{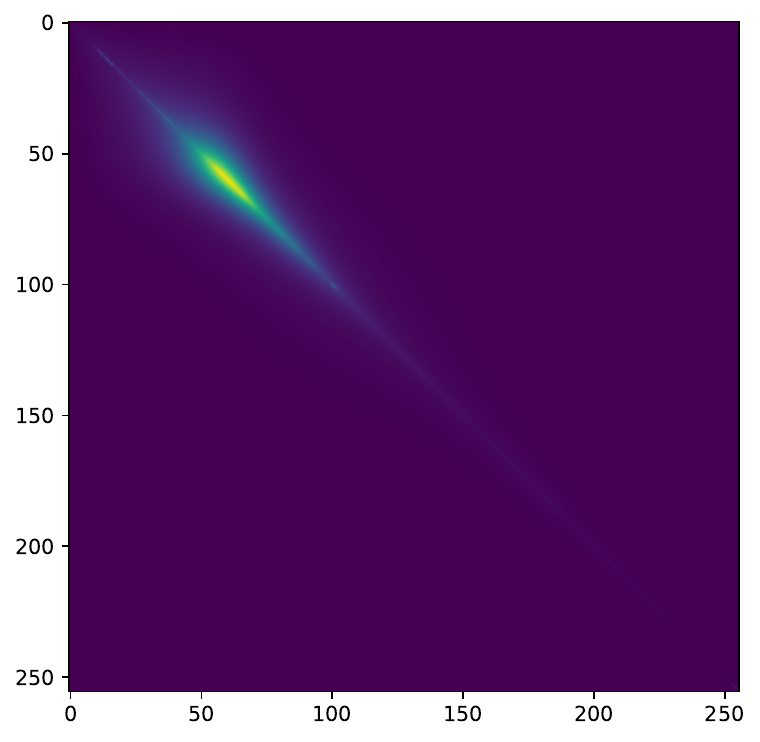}
\\
(a) TemMs & (b) SubMs & (c) TroMs & (d) TroRf \\

\end{tabular}
}
\caption{
\color{black}
The visualization of grey level co-occurrence matrix (GLCM)~\citep{1973textural} for each domain in the CASID.
}
\label{fig:glcm}
\end{center}
\end{figure}

\begin{figure}[t!]
\color{black}
\begin{center}
\includegraphics[width=0.98\linewidth]{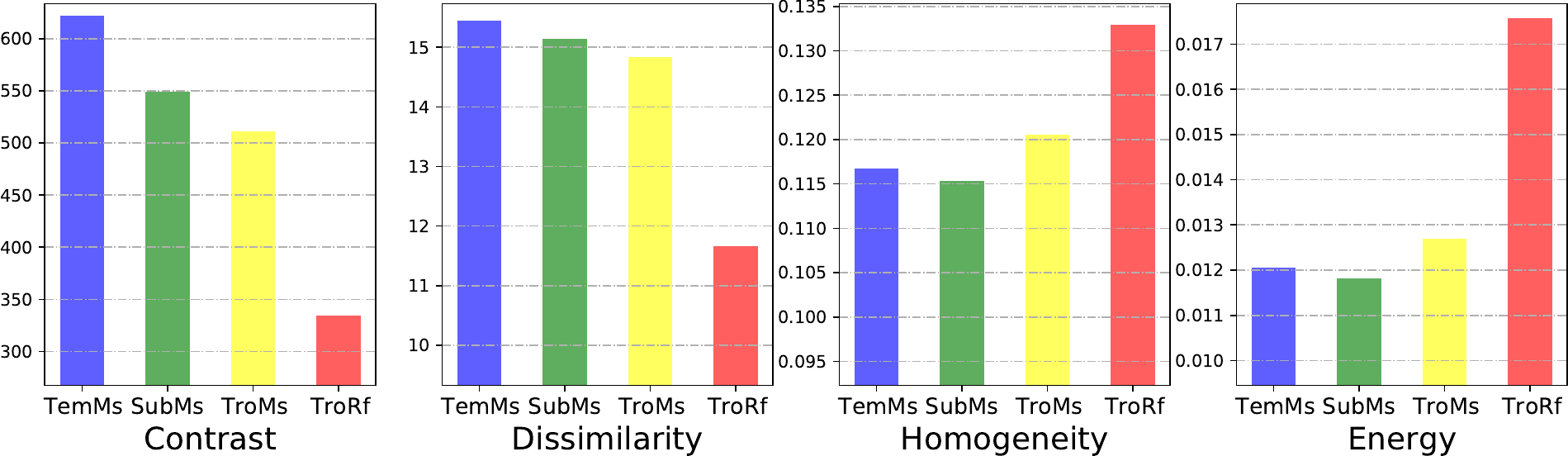}
\caption{
\color{black}
The visualization of four GLCM properties on the CASID, \ie, contrast, dissimilarity, homogeneity, and energy.
Blue: TemMs, Green: SubMs, Yellow: TroMs, Red: TroRf.
}
\label{fig:glcm_statistics}
\end{center}
\end{figure}

\begin{figure*}[t!]
\begin{center}
\scalebox{1.0}{
\begin{tabular}{cccc}
\includegraphics[width=0.98\linewidth]{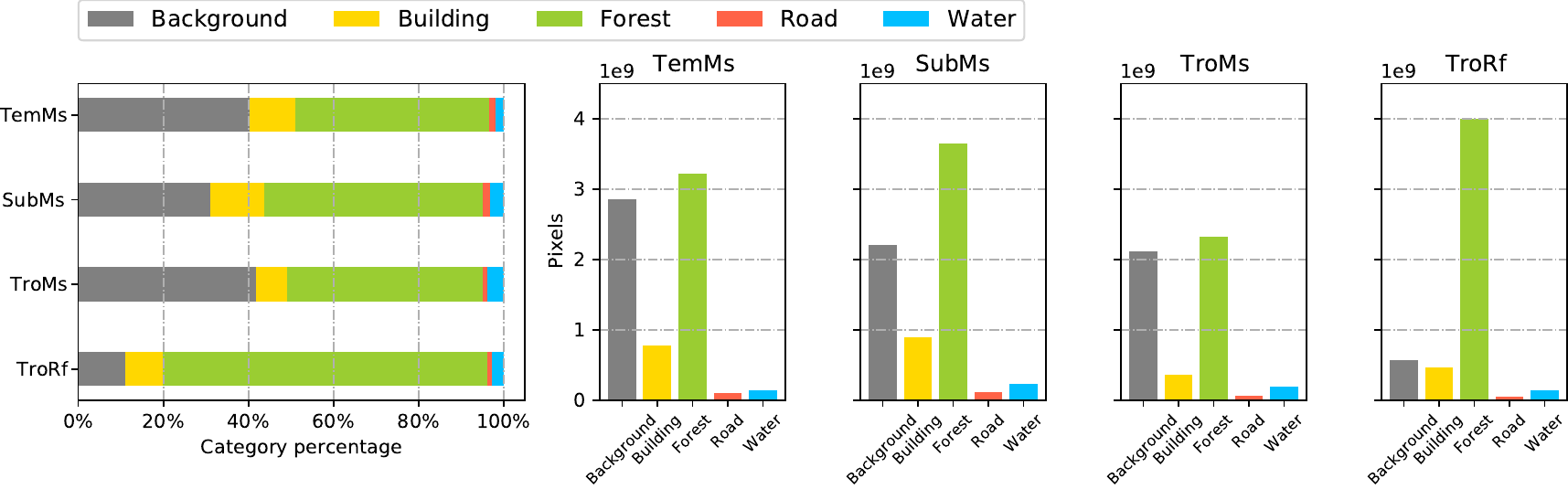} \\
\end{tabular}
}
\caption{
\color{black}
Statistics for category distributions. 
We present the category distribution and corresponding pixel sample numbers for each climate zone. 
These results demonstrate that the distribution of categories varies across climate zones.
}
\label{fig:dataset_cat}
\end{center}
\end{figure*}

\begin{figure*}[t!]
\color{black}
\begin{center}
\scalebox{0.828}{
\begin{tabular}{cccc}
\includegraphics[width=0.2918\linewidth]{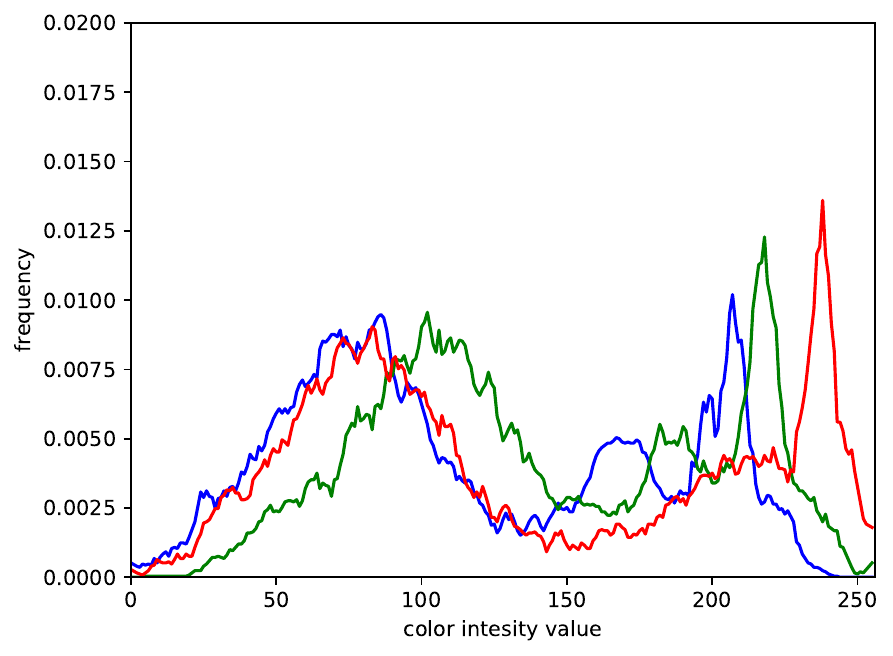}\hspace{-3mm}
&\includegraphics[width=0.2918\linewidth]{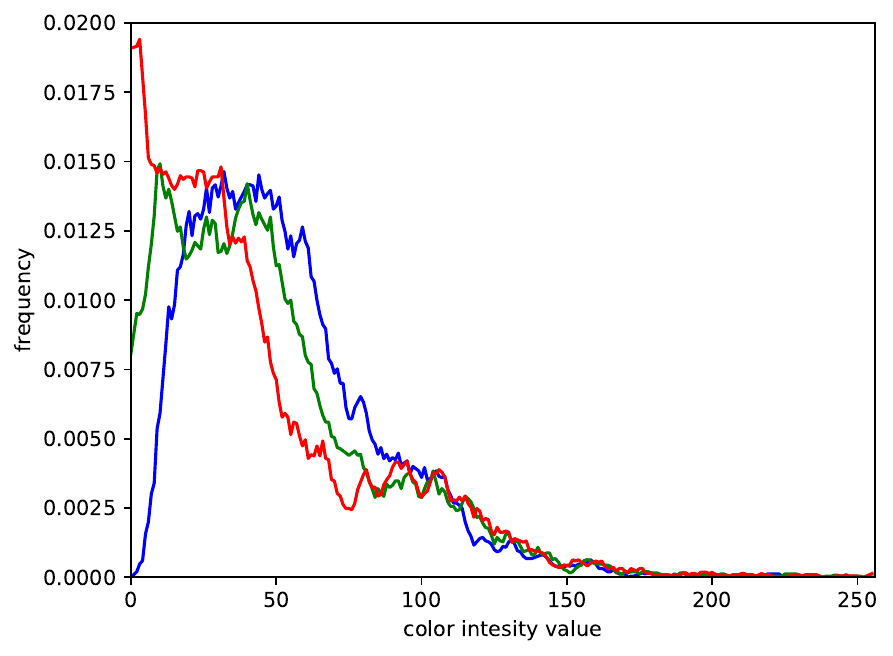}\hspace{-3mm}
&\includegraphics[width=0.2918\linewidth]{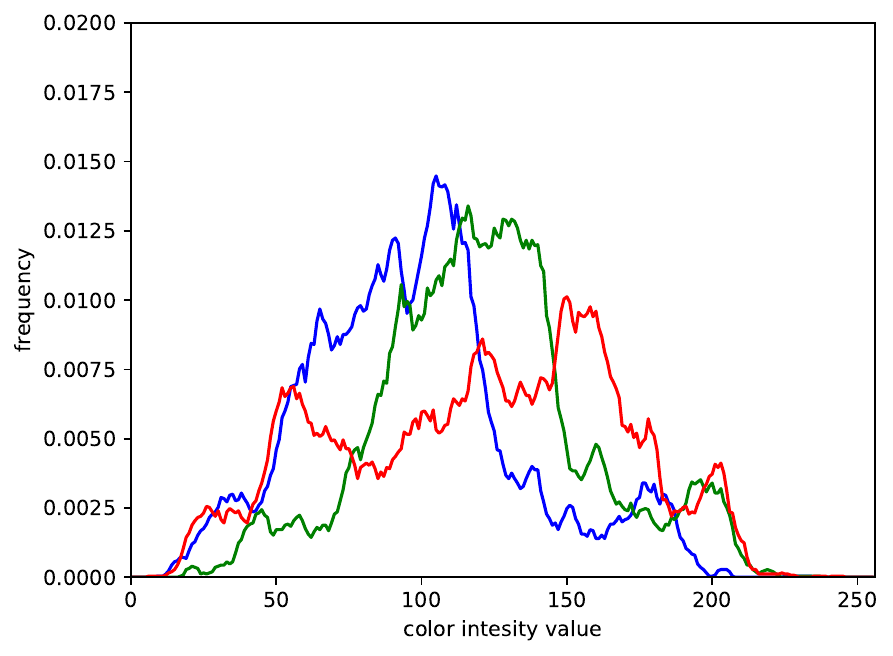}\hspace{-3mm}
&\includegraphics[width=0.2918\linewidth]{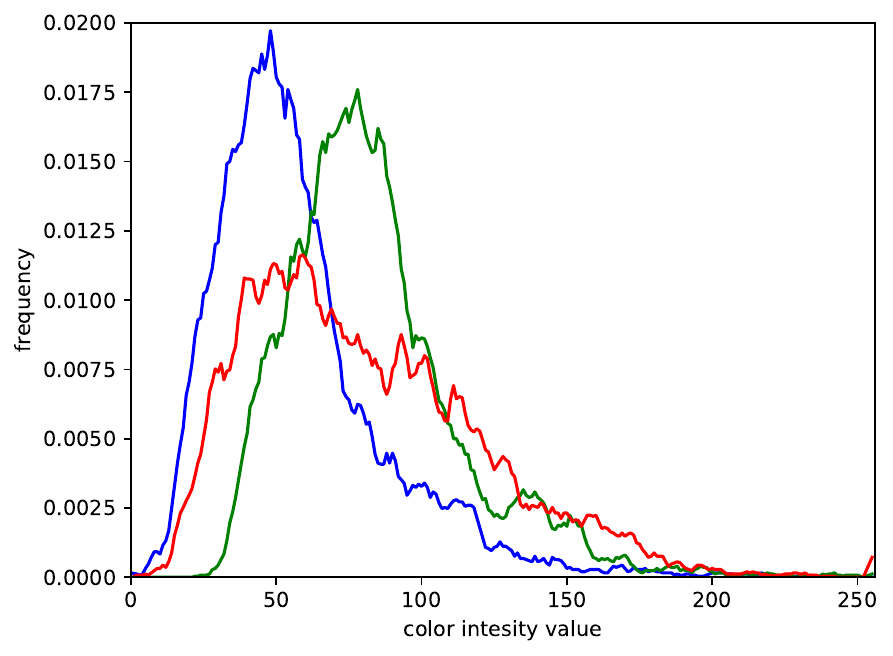}
\\
(a) TemMs & (b) SubMs & (c) TroMs & (d) TroRf \\ 
\end{tabular}
}
\caption{
\color{black}
Image coloration analysis.
We show the histograms of red, green, and blue color for each domain in the CASID.
The histogram of each domain is very different from the other domains.
This difference may contribute to the climate domain gap.
}
\label{fig:coloration_analysis}
\end{center}
\end{figure*}

\begin{figure*}[t!]
\begin{center}
\scalebox{1.0}{
\begin{tabular}{cccc}
\includegraphics[width=0.98\linewidth]{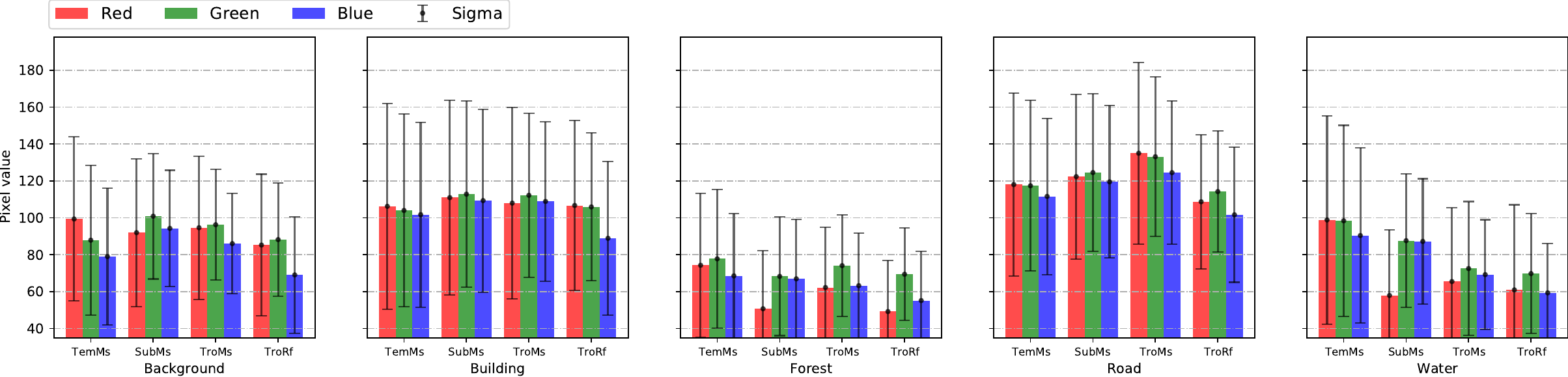} \\
\end{tabular}
}
\caption{
\color{black}
Statistics for spectral values. 
We present the mean and standard deviation (sigma) for the red, green, and blue spectral values of each category in the four climate zones. 
Our findings indicate that the spectral values of the same category vary significantly across different climate zones, resulting in large intra-class variance.
}
\label{fig:dataset_rgb}
\end{center}
\end{figure*}

\begin{figure}[t!]
\color{black}
\begin{center}
\includegraphics[width=0.98\linewidth]{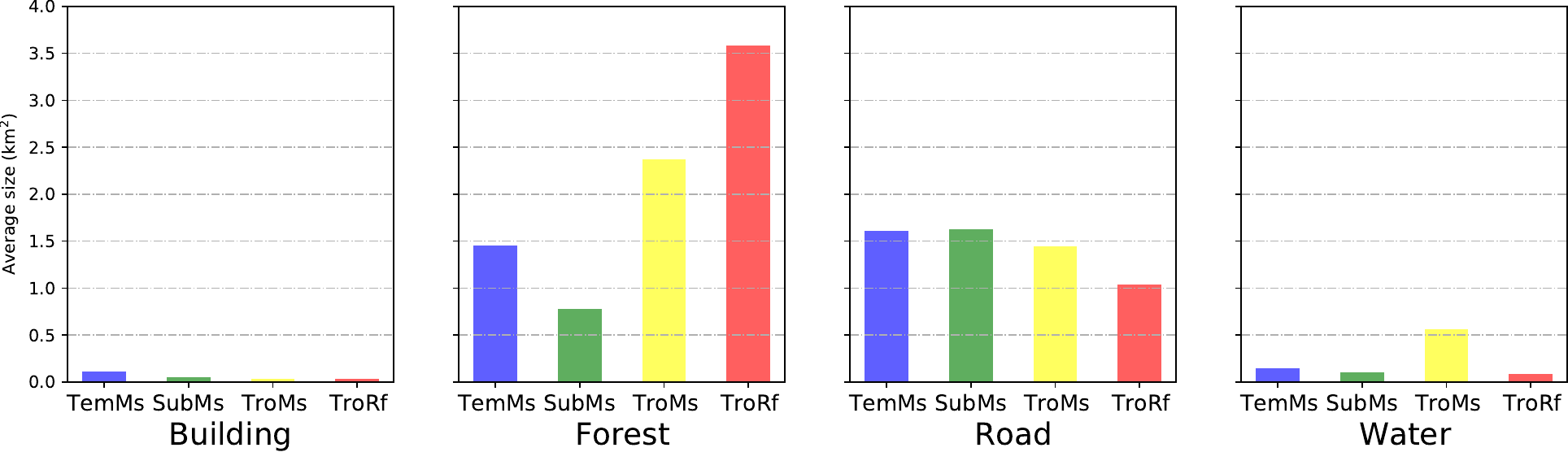}\\
\includegraphics[width=0.98\linewidth]{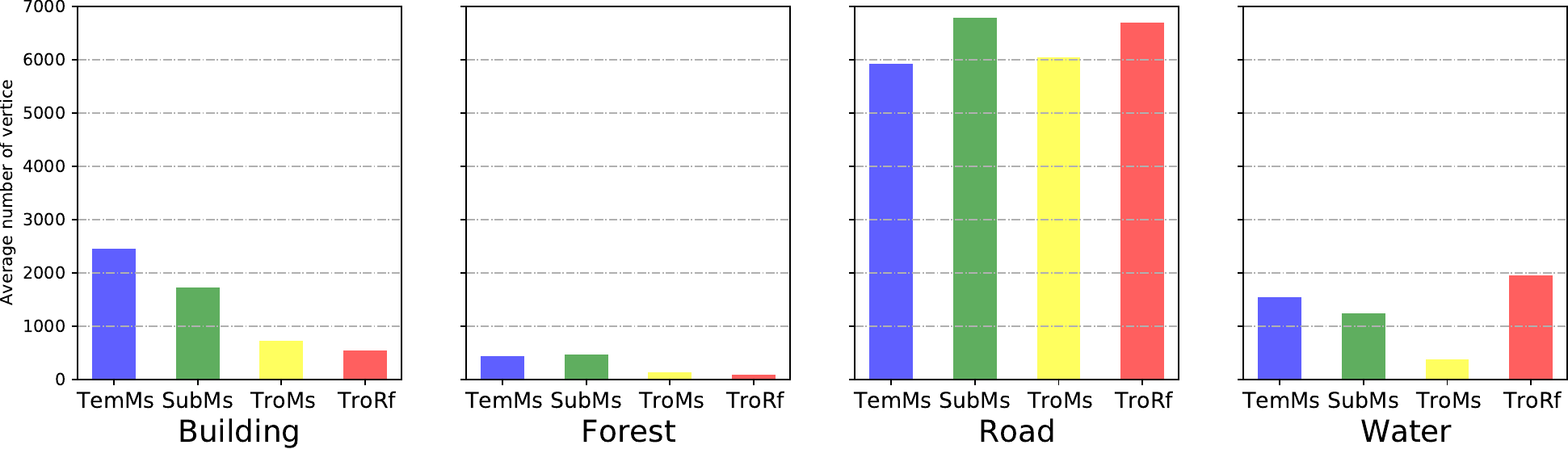}
\caption{
\color{black}
Statistics of object shape. 
The average size depicts the scale of objects, and the number of vertices in an object polygon represents the complexity of its shape.
}
\label{fig:average_object_size_vertice}
\end{center}
\end{figure}

\begin{figure}[t!]
\color{black}
\begin{center}
\includegraphics[width=0.98\linewidth]{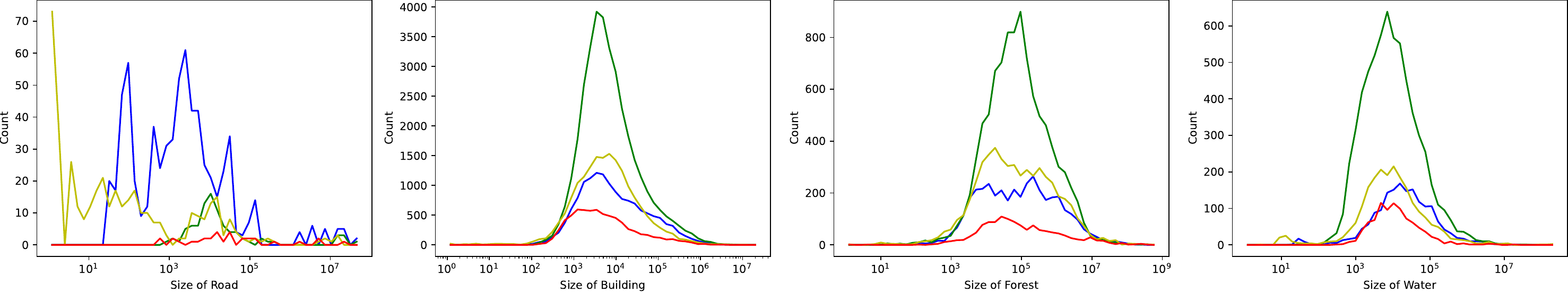}\\
\includegraphics[width=0.98\linewidth]{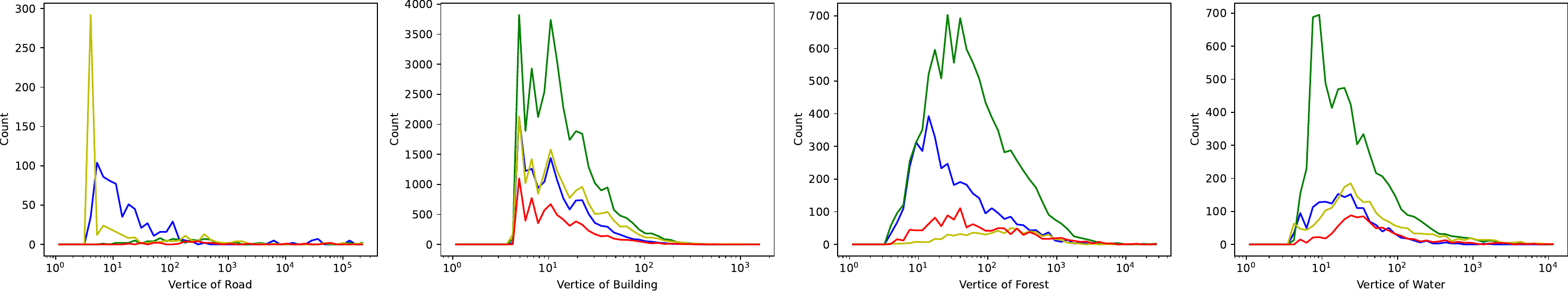}
\caption{
\color{black}
Histogram of object shape distribution. 
The size depicts the scale of objects, and the vertice represents the complexity of its shape.
}
\label{fig:object_size_vertice}
\end{center}
\end{figure}

\begin{table*}[t]
\color{black}
\begin{center}
\vspace{-0mm}
\caption{
\color{black}
Semantic segmentation results on the test set of CASID.
We conduct all experiments five times and present the mean and variance results below.
Bkgd: Background, Bldg: Building, Frst: Forest, Rd: Road, Wtr: Water.
}
\label{tab:ss}
\vspace{-0mm}
\scalebox{0.918}{
\begin{tabular}{l|c|ccccc|c|cc}
\toprule[1.5pt]
\multirow{2}{*}{Methods} & \multirow{2}{*}{Backbone} 
& \multicolumn{5}{c|}{IoU (\%)} & \multirow{2}{*}{mIoU} & \multirow{2}{*}{Dice} \\
\cline{3-7}
& 
&Bkgd &Bldg &Frst &Rd &Wtr & \\
\midrule
FCN8S \citep{fcn2015} & VGG-16
 & 51.9 & 75.5 & 74.7 & 32.2 & 51.6 & 57.2 {\color{gray}($\pm$0.3)} & 71.3 {\color{gray}($\pm$0.3)} \\ 
U-Net \citep{unet} & ResNet-50  
 & 53.3 & 77.5 & 76.3 & 41.0 & 60.9 & 61.8 {\color{gray}($\pm$1.2)}& 75.6 {\color{gray}($\pm$1.0)}\\ 
PSPNet \citep{pspnet} & ResNet-50  
 & 53.7 & 77.1 & 75.3 & 34.0 & 45.3 & 57.1 {\color{gray}($\pm$0.4)}& 71.2 {\color{gray}($\pm$0.3)}\\ 
LinkNet \citep{linknet} & ResNet-50  
 & 52.8 & 78.0 & 76.8 & 40.8 & 63.4 & 62.3 {\color{gray}($\pm$0.2)}& 75.8 {\color{gray}($\pm$0.2)}\\ 
DeepLabV3+ \citep{deeplabv3plus} & ResNet-50  
 & 53.4 & 76.2 & 74.5 & 38.9 & 42.5 & 56.9 {\color{gray}($\pm$0.5)} & 71.4 {\color{gray}($\pm$0.6)}\\ 
PAN \citep{2018pan} & ResNet-50  
 & 54.9 & 76.9 & 75.6 & 35.7 & 37.2 & 56.0 {\color{gray}($\pm$0.7)}& 70.1 {\color{gray}($\pm$0.6)}\\ 
U-Net++ \citep{unet++} & ResNet-50  
 & 53.2 & 77.8 & 76.2 & 42.3 & 60.0 & 61.9 {\color{gray}($\pm$0.6)}& 75.6 {\color{gray}($\pm$0.4)}\\
HRNet \citep{hrnet} & HRNetW32 
 & 52.5 & 78.6 & 74.4 & 43.8 & 55.2 & 60.9 {\color{gray}($\pm$0.8)}& 74.8 {\color{gray}($\pm$0.6)}\\ 
FarSeg \citep{2020foreground} & ResNet-50  
 & 52.5 & 77.0 & 73.8 & 40.8 & 46.6 & 58.2 {\color{gray}($\pm$0.7)}& 72.5 {\color{gray}($\pm$0.7)}\\ 
FactSeg \citep{factseg} & ResNet-50  
 & 52.8 & 77.5 & 74.6 & 38.9 & 46.1 & 58.0 {\color{gray}($\pm$1.1)}& 72.2 {\color{gray}($\pm$1.0)}\\ 
SegFormer~\citep{2021segformer} & MiT-B0
 & 55.6 & 77.5 & 77.0 & 36.9 & 60.7 & 61.7 {\color{gray}($\pm$2.0)}& 75.0 {\color{gray}($\pm$1.5)}\\ 
SegFormer~\citep{2021segformer} & MiT-B5
 & 54.9 & 78.5 & 76.1 & 42.7 & 62.8 & 63.0 {\color{gray}($\pm$0.2)}& 76.4 {\color{gray}($\pm$0.1)}\\ 
PoolFormer~\citep{poolformer} & PoolFormer-S12
& 55.0 & 77.1 & 76.5 & 36.5 & 57.8 & 60.6 {\color{gray}($\pm$1.6)} & 74.3 {\color{gray}($\pm$1.3)} \\ 
PoolFormer~\citep{poolformer} & PoolFormer-M48 
& 55.8 & 77.7 & 77.5 & 38.6 &\bf 63.8 & 62.7 {\color{gray}($\pm$2.2)}& 75.9 {\color{gray}($\pm$1.7)} \\ 
SegNeXt~\citep{segnext} & MSCAN-T
 &\bf 57.0 & 78.2 &\bf 78.3 & 39.9 & 50.3 & 60.7 {\color{gray}($\pm$0.8)}& 74.4 {\color{gray}($\pm$0.7)} \\ 
SegNeXt~\citep{segnext} & MSCAN-L
 &\bf 57.0 &\bf 79.2 & 77.9 &\bf 44.1 & 58.6 &\bf 63.4 {\color{gray}($\pm$1.1)}&\bf 76.7 {\color{gray}($\pm$0.9)} \\
\bottomrule[1.5pt]
\end{tabular}}
\end{center}
\vspace{-5mm}
\end{table*}

\begin{figure*}[t]
\begin{center}
\scalebox{1.0}{
\begin{tabular}{cccccc}
\hspace{-3mm}
\includegraphics[width=0.98\linewidth]{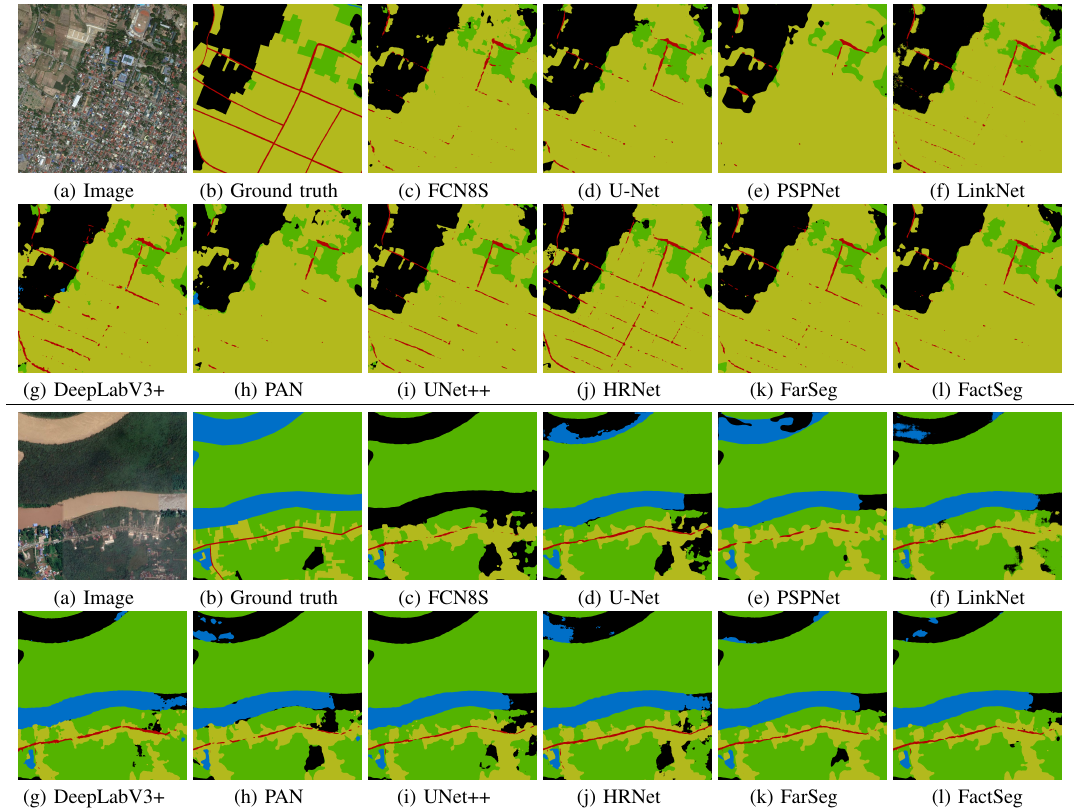}
\end{tabular}
}
\caption{
Representative visualized segmentation results.
The top two rows are the samples from TemMs set, and the bottom rows are the samples from SubMs set.
}
\vspace{-2.8mm}
\label{fig:seg1}
\end{center}
\end{figure*}

\begin{figure*}[t]
\begin{center}
\scalebox{1.0}{
\begin{tabular}{cccccc}
\hspace{-3mm}
\includegraphics[width=0.98\linewidth]{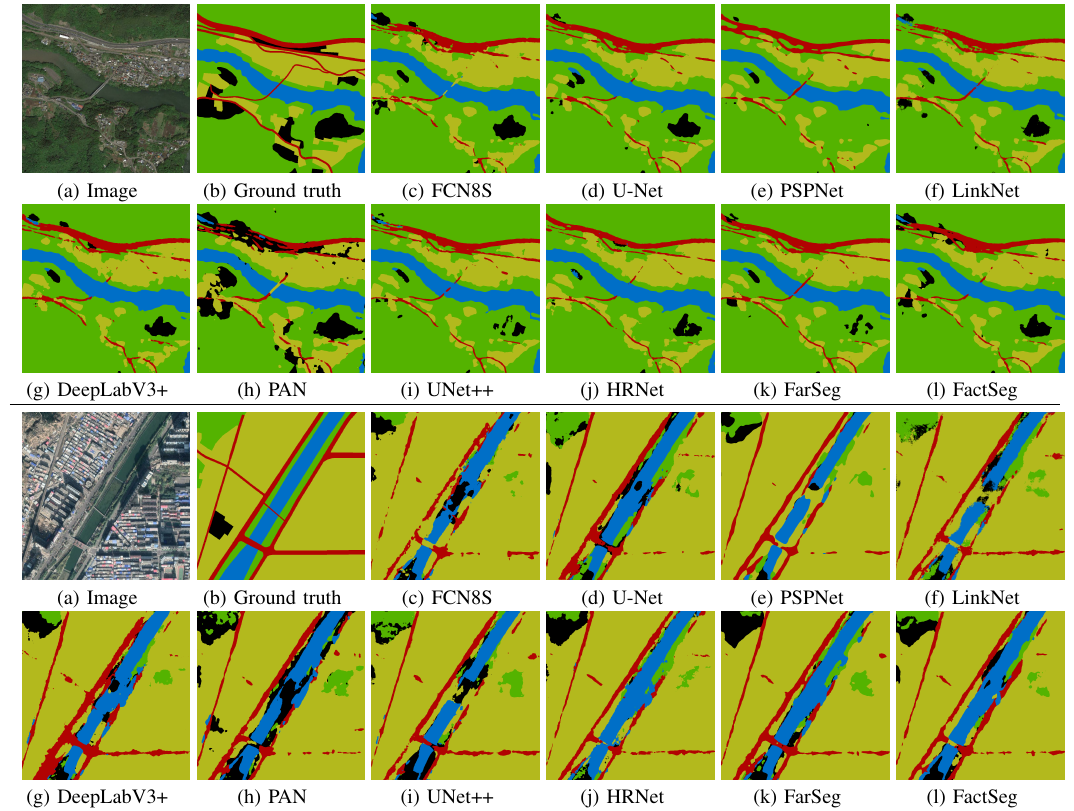}
\end{tabular}
}
\caption{
Representative visualized segmentation results.
The top two rows are the samples from TroMs set, and the bottom rows are the samples from TroRf set.
}
\label{fig:seg2}
\end{center}
\end{figure*}

\begin{figure*}[t]
\begin{center}
\scalebox{1.0}{
\begin{tabular}{cccccc}
\hspace{-3mm}
\includegraphics[width=0.98\linewidth]{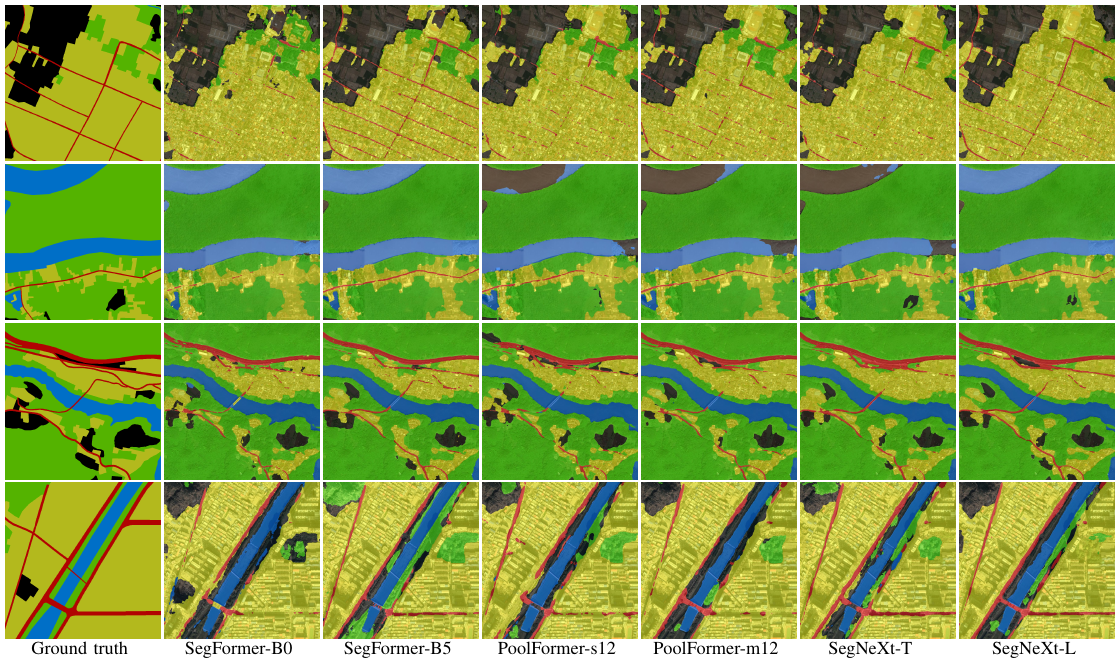}
\end{tabular}
}
\caption{
\color{black}
Representative visualized segmentation results.
From top to bottom are the samples from TemMs, SubMs, TroMs, and TroRf set.
}
\label{fig:seg3}
\end{center}
\end{figure*}

\section{Related works}
\label{sec:related_works}

In this section, we first describe the development of land-cover datasets for semantic segmentation and then introduce semantic segmentation methods. 
{
Finally, we provide an overview of unsupervised domain adaptation methods.
}

\subsection{Land-Cover Datasets for Semantic Segmentation}

Recently, with the advancement of deep learning technology, large-scale datasets have played a crucial role in advancing scene understanding tasks such as land-cover semantic segmentation~\citep{2020gid, 2021loveda}.
As shown in Table~\ref{tab:datasets}, datasets such as ISPRS Vaihingen, ISPRS Potsdam, Zurich Summer~\citep{2015Zurich}, Zeebruges~\citep{2018Zeebruges}, USGS SIRI-WHU~\citep{2015usgs}, EvLab-SS~\citep{2017evlab}, and LandCover.ai~\citep{2021landcover} are collected for land-cover semantic segmentation.
These datasets provide remote sensing images with accurate pixel-wise land-cover labels and have contributed to the development of land-cover semantic segmentation methods.
However, these datasets cover relatively small areas and contain a limited number of annotated image samples, which limits the generalization ability of relevant algorithms.

{
To address this limitation, several large-scale datasets have been proposed, including SpaceNet~\citep{2018spacenet}, DeepGlobe~\citep{2018deepglobe}, AIRS~\citep{2019AIRS}, LandCoverNet~\citep{2020landcovernet}, GID dataset~\citep{2020gid}, and LoveDA~\citep{2021loveda}.
While these datasets collect image samples from different cities, they do not adequately consider the domain gap caused by the significant diversity among different geographic areas.

Specifically, land cover can be influenced and diversified by two factors: human management and climate~\citep{2021loveda, dale1997relationship, reidmiller2019fourth}.
LoveDA~\citep{2021loveda} demonstrates that human management contributes to differentiation in the human living environment due to societal and economic development, causing a domain gap between urban and rural zones for land-cover semantic segmentation.
OpenEarthMap~\citep{2023openearthmap} collects samples from six continents to construct a global benchmark dataset with 2.2 million segments of aerial and satellite samples with large image sizes.
However, the impact of climate on land-cover semantic segmentation, which is more profound than human management, has received limited investigation.
The domain gap caused by climate can significantly degrade the performance of semantic segmentation methods, rendering them ineffective in practical applications.
To improve the model's generalizability for large-scale land-cover semantic segmentation, we have collected and annotated a large-scale Climate-Aware Satellite Image Dataset for land-cover semantic segmentation.
}

\subsection{Semantic Segmentation}
{
Land-cover semantic segmentation is a long-standing research topic that assigns a label to each pixel, also known as pixel-level classification. 
As a powerful technique for Earth observation, it has been widely explored over the past few decades
}
Existing satellite image semantic segmentation methods are eminently motivated by the fully-convolutional networks (FCNs) \citep{fcn2015}.
Several earlier works \citep{2017frommaps, 2018classwithedge, 2018landcovermapping} focus on using multi-level semantic features on local patterns of images using deep CNN. 
Besides, there are also some methods \citep{2019relation} capturing long-range spatial relationships between entities by proposing spatial and channel relation modules. 
Though previous works have achieved great advances, there are two critical issues, one is an extremely foreground-background imbalanced distribution, and the other is multiple small objects along with the complex background. 
Recently, FarSeg~\citep{2020foreground} utilizes relation-based and optimization-based foreground modeling to handle the foreground-background imbalance problems in remote sensing imagery. 
Factseg~\citep{factseg} uses a foreground activation object representation framework made up of a dual-branch decoder and collaborative probability loss. It effectively suppresses background distractions while also enhancing the small object features.

\subsection{Unsupervised Domain Adaptation}

Unsupervised domain adaptation is a technique aimed at reducing the performance loss caused by the domain gap. 
{
It involves transferring a model trained on the source domain to the target domain. 
This technique is effective in reducing the domain gap caused by climate. 
Recent advancements in unsupervised domain adaptation have been made in two branches: adversarial training and self-training.
}

{
Adversarial training methods typically involve an extractor that learns domain-invariant features and a discriminator that distinguishes these features. 
}
Tsai~\etal~\citep{adaptseg} utilize spatial similarities in semantic outputs to adapt the structured output space through adversarial learning. 
To address the misalignment of category-level joint distribution, 
Luo~\etal~\citep{CLAN} introduce a category-level adversarial network that aligns each class with an adaptive adversarial loss. 
Wang~\etal~\citep{FADA} align class-level features through a fine-grained adversarial learning framework. 
To avoid distorting the target-domain feature distribution, Ning~\etal~\citep{MADA} incorporate active learning to assist domain adaptation and achieve superior segmentation performance.

Self-training methods generate pseudo-labels on the target data and fine-tune the model on the source data alternately.
Lian~\etal~\citep{PyCDA} connect the self-training and curriculum adaptation and introduces the self-motivated pyramid curriculum.
It observes the target properties and fuses multi-scale features for semantic segmentation domain adaptation. 
To solve the class imbalance problem of pseudo-labels in self-training, Zou~\etal~\citep{CBST} propose a class-balanced self-training strategy to sample pseudo-labels.
To improve the quality of pseudo-labels and achieve sample balance, Mei~\etal~\citep{IAST} design an instance adaptive self-training selector, which can involve more useful information for training.
Recently, Wang~\etal~\citep{wang2021uncertainty} propose a novel uncertainty-aware pseudo label refinery framework for utilizing both accurate prediction and uncertain prediction with low confidence in pseudo labels.
Cheng~\etal~\citep{cheng2021dualpath} propose a novel dual-path learning framework to alleviate visual inconsistency.
Zhou~\etal~\citep{2020ASA} leverage co-occurring patterns between pairwise pixels in the output of structured semantic segmentation.
Yuan~\etal~\citep{2022BAFFT} propose an Inter-class Separation and Intra-class Aggregation (ISIA) mechanism to address the class confusion problem and designs an Adaptive-weighted Instance Matching (AIM) strategy for instance-level adaptation.
Hoyer~\etal~\citep{2022daformer} propose DAFormer, which utilizes a Transformer encoder and multi-level context-aware feature fusion decoder and achieves state-of-the-art performance with significant improvement.

For the remote sensing community, there have been a series of works exploring few-shot transfer learning for Synthetic Aperture Radar (SAR) image classification~\citep{2022fewshotsar, 2019afastinference, 2019deeptransfer, 2022afewshowlearning}.
Rostami~\etal~\citep{2019deeptransfer} propose a framework for SAR image classification that transfers knowledge from the Electro-Optical (EO) domain to eliminate the need for a large labeled dataset.
Tang~\etal~\citep{2019afastinference} utilize an improved Siamese Network approach for SAR target few-shot learning.
Gao~\etal~\citep{2022afewshowlearning} introduce a novel few-shot learning method for SAR image recognition, which consists of a multi-feature fusion network (MFFN) and a weighted distance classifier (WDC) to enhance the feature extraction and classification capability.
Tai~\etal~\citep{2022fewshotsar} propose the connection-free attention module to selectively transfer features shared between EO and SAR samples from a source network to a target network to supplement the loss of information brought by extra SAR samples. 
Besides, unsupervised domain adaptation methods also have been studied for land-cover semantic segmentation~\citep{2020gid, 2020standardgan}.
These methods follow the general unsupervised domain adaptation approach in the computer vision field, with some improvements.
However, due to the lack of datasets with consistent annotation granularity and high diversity, the advancement of unsupervised domain adaptation algorithms for remote sensing has been limited.
To push the advancements further, the CASID dataset is proposed as a more challenging benchmark, promoting future research of remote sensing unsupervised domain adaptation algorithms and relevant applications.


\section{Dataset Description}
\label{sec:dataset}
{\color{black}
This section first describes the characteristics of the proposed Climate-Aware Satellite Image Dataset (CASID) for domain adaptive land-cover semantic segmentation. Then, it provides a statistical analysis of the climate domain gap.
}


\subsection{Overview and Dataset Properties}

{
The climate factor has a significant impact on land cover features, posing a challenge to existing deep learning models.
In order to advance satellite imagery for earth observation and enhance generalizability, we are investigating the impact of climate on land-cover semantic segmentation for the first time.
We introduce a large-scale Climate-Aware Satellite Image Dataset for land-cover semantic segmentation, named CASID, which can be utilized for two challenging tasks: land-cover semantic segmentation and unsupervised domain adaptation.
The main characteristics of this dataset can be summarized as follows:
} 


\vspace{+1mm}
\noindent{\bf Climate-aware.}
To investigate the impact of climate on the land-cover semantic segmentation, we collect satellite images from four typical climate zones, \ie, temperate monsoon, subtropical monsoon, tropical monsoon, and tropical rainforest~\citep{1980worldclimate}. 
Samples from four typical climate zones form four sub-datasets/domains regarding their climate types.
Owing to different temperatures and precipitation, satellite images from different climate zones show different manifestations, as shown in Figure~\ref{fig:CASID}.
We also statistically analyze the images from these four typical climate zones and find their features strongly correlate with climate in Section~\ref{sec:climate_domain_gap}.

\vspace{+1mm}
\noindent{\bf Large-scale, high diversity, and large image size. }
{
To increase the diversity of the CASID dataset, we collect 980 high-quality satellite images with a size of $5,000\times 5,000$ pixels obtained from the Google Earth platform, whose spatial resolution is $1$ meter.
}
They are acquired from 30 different cities/regions around Asia, including densely populated areas and sparsely populated suburbs.
These cities have very different land cover features.
They are widely distributed over geographic areas covering more than 24,500 km$^2$ and have very different land cover features, as shown in Figure~\ref{fig:CASID}.
Due to the extensive geographical distribution, CASID represents diversified distribution information of ground objects in different areas.
And to keep the sample balance, we collect a similar number (200$\sim$300) of image samples for each climate.
As shown in Table~\ref{tab:cities}, for each type of climate, we collect samples from 6$\sim$10 cities/regions, and 2 of them are kept as the test set for evaluation.
In this way, we split $\sim$70\% of these images into the train set and leave the rest of $\sim$30\% as the test set.

\vspace{+1mm}
\noindent{\bf{High inter- and intra-domain variance.}}
{\color{black}
Thanks to its large-scale, high diversity, and climate-aware properties, CASID exhibits high inter- and intra-domain variance, as shown in the t-SNE visualization in Section~\ref{sec:climate_domain_gap}. 
The high inter-domain variance makes it more challenging, as demonstrated in Table~\ref{tab:crossdomain}, while the high intra-domain variance benefits unsupervised domain adaptation, as discussed in Section~\ref{sec:uda}.
}

\vspace{+1mm}
\noindent{\bf{Well annotated. }}
For each satellite image, we provide accurate pixel-wise semantic segmentation labels for five categories, \ie, background, building, forest, road, and water. 
They are annotated by a group of experts and labeled with black, yellow, green, red, and blue, as shown in Figure~\ref{fig:CASID}.

\vspace{+1mm}
\noindent{\bf{Complex background}. }
The remote sensing land-cover semantic segmentation task is always faced with complex background samples (\ie, land-cover objects that are not of interest)~\citep{pang2019mathcal, 2020foreground}.
{
The large image size and diverse scenes in the CASID dataset provide rich details and introduce a larger intra-class variance for the background samples. 
This benefits the learning of semantic segmentation models and helps them effectively handle the complex background in remote sensing images.
}

\vspace{+1mm}
\noindent{\bf{Category imbalance}.}
{
Different land types naturally exhibit diverse shapes and sizes. 
For instance, the road category is characterized by its elongated and narrow form, while the forest category can span a significant surface area. 
This inherent variation in shape and size contributes to the challenge of category imbalance. As a result, the proportions of different land categories differ significantly, further exacerbating the issue. 
The presence of category imbalance poses a unique challenge for tasks such as semantic segmentation and unsupervised domain adaptation. Addressing this challenge requires developing effective strategies to handle the varying proportions of land categories within the dataset.
}

\subsection{Statistics Analysis for Climate Domain Gap}
\label{sec:climate_domain_gap}
The climate greatly affects the manifestation of land cover. 
In general, the climate is the major determinant of vegetation~\citep{prentice1990bioclimatic, woodward2004global, salazar2007climate}.
Climate zones with higher temperatures and more rainfall support more productive and diverse plant communities, while regions with lower temperatures and less precipitation typically have less productive vegetation. 
The climate also impacts other land cover types, such as water bodies, which are affected by precipitation and evapotranspiration.

Additionally, the appearance of buildings and roads are also affected by the climate, they are designed with considerations of temperature and precipitation~\citep{golany1996urban,givoni1998climate}. 
{
For example, areas with high precipitation levels may require the use of different building materials or construction methods to prevent water damage. 
Areas with high temperatures may require materials with higher thermal insulation to maintain a comfortable indoor temperature. 
Similarly, roads may be designed with different materials and widths based on the climate and weather patterns in the region. 
These variations in building and road design can result in significant differences in their spectral values, further contributing to the climate domain gap for land-cover semantic segmentation.
} 
As shown in Figures~\ref{fig:CASID} and~\ref{fig:dataset_tSNE}, the satellite images from different climates zone show different perceptual characteristics to human eyes.
{
We refer to these differences caused by climate as the climate domain gap. 
To statistically describe the climate domain gap, we further analyze it in various aspects, including global image features, image texture, category distribution, spectral values, and object shape.
}

\vspace{+1mm}
\noindent{\bf{Global image feature.}}
{
To analyze the differences in global image features between images from different climate zones, we first uniformly sample images from the CASID dataset and crop them into $1024\times 1024$ image patches. 
Then, following the approach in~\citep{IS}, we employ InceptionV3~\citep{InceptionV3} to generate a global pooled 2048-dimensional vector as the global image feature representation for each image patch. 
Finally, we utilize t-SNE~\citep{2008tsne} for dimension reduction and visualization.
}
The two-dimensional t-SNE visualization of the CASID dataset is shown in Figure~\ref{fig:dataset_tSNE}. 
{
The images from the four climate zones are clearly separated by t-SNE, indicating a significant domain gap caused by climate. } 
Interestingly, the t-SNE visualization in Figure~\ref{fig:dataset_tSNE} exhibits a similar pattern to the distribution of climate areas on the climate map in Figure~\ref{fig:CASID}.

{
To provide further elaboration, we present additional visualized t-SNE results in Figure~\ref{fig:climate_tSNE}. 
In Figure~\ref{fig:climate_tSNE}(a), a clear boundary and distinct distributions can be observed between the TemMs domain and the TroRf domain. 
This indicates a significant domain gap between these two domains. 
When we introduce samples from the SubMs domain or the TroMs domain, as shown in Figures~\ref{fig:climate_tSNE}(b) and (c), most of these samples are positioned between the TemMs domain and the TroRf domain. 
This suggests that the domain differences between SubMs/TroMs and TemMs/TroRf are less pronounced compared to the difference between TemMs and TroRf. 
It demonstrates a gradual change in image features from TemMs to TroRf. Therefore, the t-SNE result of the four domains in Figure~\ref{fig:climate_tSNE}(d) is consistent with the geographical location differences between these domains.
These results suggest that the climate zone is a significant factor in influencing image manifestation and global image features.
}

{\color{black}
\vspace{+1mm}
\noindent{\bf{Image texture. }}
We use a classic grey level co-occurrence matrix (GLCM)~\citep{1973textural} for texture feature extraction.
As shown in Figure~\ref{fig:glcm}, we can see the visualization of GLCM for each domain in the CASID gradually change from left (TemMs) to right (TroRf), which is consistent with the global image feature analysis.
We further calculate the texture properties of GLCM, \ie, contrast, dissimilarity, homogeneity, and energy.
These properties are computed as follows:
\begin{equation}
\begin{aligned}
\text{Contrast}=\sum_{i=0}^{N-1}\sum_{j=0}^{N-1}(i-j)^2\cdot P_{i,j}.
\end{aligned}
\end{equation}
\begin{equation}
\begin{aligned}
\text{Dissimilarity}=\sum_{i=0}^{N-1}\sum_{j=0}^{N-1}|i-j|\cdot P_{i,j}.
\end{aligned}
\end{equation}
\begin{equation}
\begin{aligned}
\text{Homogeneity}=\sum_{i=0}^{N-1}\sum_{j=0}^{N-1}\frac{P_{i,j}}{1+(i-j)^2}.
\end{aligned}
\end{equation}
\begin{equation}
\begin{aligned}
\text{Energy}=\sum_{i=0}^{N-1}\sum_{j=0}^{N-1} P_{i,j}^2.
\end{aligned}
\end{equation}
where $(i,j)$ are the coordinates of the GLCM, $N$ indicates grey level, and $P_{i,j}$ represents the probability of $(i,j)$ in the GLCM.

Contrast measures the local variations in intensity.
Dissimilarity measures the dissimilarity between adjacent pixels in the image.
Homogeneity measures the similarity or uniformity of neighboring pixel values. 
Energy measures the magnitude or strength of the texture in the image.
As shown in Figure~\ref{fig:glcm_statistics}, statistical analysis on the basis of GLCM demonstrates that contrast, dissimilarity, homogeneity, and energy also gradually change from TemMs to the TroRf domain.
These results show the texture of images is associated with the domain gap and correlated with the climate zone.
}

\vspace{+1mm}
\noindent{\bf{Category distribution. }}
{
As shown in Figure~\ref{fig:dataset_cat}, different climate zones show significant variations in class distributions. 
Due to variances in temperature and precipitation, regions within the tropical rainforest climate zone are typically characterized by dense rainforests, resulting in a substantial proportion of the forest category, specifically 76.2\%. 
In contrast, the temperate monsoon climate zone generally features less forest coverage, with the percentage of forest pixels being considerably lower than that of the tropical rainforest climate zone (76.2\% for tropical rainforest area \emph{vs}. 45.4\% for temperate monsoon area). 
This observation underscores the impact of climate on the distribution of land-cover categories.
}

\vspace{+1mm}
\noindent{\bf{Spectral value. }}
{
Figure~\ref{fig:coloration_analysis} illustrates histograms of the red, green, and blue for each domain in the CASID dataset. 
Each domain possesses a distinct histogram that reflects its unique characteristics.
In Figure~\ref{fig:dataset_rgb}, we further compare the spectral value distributions of different categories across four climate zones. 
Notably, the spectral values of the same category vary across the different climates. 
For instance, the forest category in the tropical rainforest area demonstrates lower mean values for red and blue channels. 
On the other hand, the background in the temperate monsoon area exhibits a higher mean value of red, indicating significant inter-class variance among the different climate zones.
}

{\color{black}
\vspace{+1mm}
\noindent{\bf{Object shape}.}
We present the object shape statistics in Figure~\ref{fig:average_object_size_vertice}, where the average size represents the scale of objects, and the number of vertices in an object polygon reflects the complexity of its shape. 
Different categories of land types exhibit various shapes and sizes. 
For example, the road category is thin, long, and curved, resulting in relatively more vertices. 
On the other hand, forests can cover large surface areas, leading to a relatively large average size. 
Moreover, the average object size and number of vertices in each category vary from TemMs to TroRf, indicating that object shapes of land-cover categories are associated with the climate. Additional histograms of object shape distribution for each category are shown in Figure~\ref{fig:object_size_vertice}.
}

\begin{table}[t!]
\color{black}
\caption{
\color{black}
Results reported on LoveDA~\citep{2021loveda} with semantic segmentation model~\citep{deeplabv2}.
It is trained and evaluated on images from different domains.
}
\begin{center}
\scalebox{0.8}{
\begin{tabular}{l|c|cc}
\toprule[1.5pt]
\diagbox{Train on}{mIoU}{Test on} & Rural set & Urban set \\
\midrule
Rural set  & \bf 46.8 & 31.3 \\
Urban set  & 31.7 &\bf 45.7 \\
\midrule
\it Performance gap & 15.1 & 14.4 \\
\bottomrule[1.5pt]
\end{tabular}
}
\end{center}
\label{tab:loveda_crossdomain}
\end{table}

\begin{table}[t!]
\color{black}
\caption{
\color{black}
Results reported on OpenEarthMap~\citep{2023openearthmap} with semantic segmentation model~\citep{deeplabv2}.
It is trained and evaluated on images from different domains.
Asia: AS, Europe: EU, Africa: AF, North America: NA, South America: SA, and Oceania: OC.
}
\vspace{-2.8mm}
\begin{center}
\scalebox{0.8}{
\begin{tabular}{l|c|c|c|c|c|c}
\toprule[1.5pt]
\diagbox{Train on}{mIoU}{Test on} &AF &AS &EU &NA &OC &SA  \\
\midrule
AF &\bf 49.5 &35.2 & 27.5  & 31.8 &35.6 &36.9 \\
AS & 35.9 &\bf 52.4 &33.8 &34.6 &46.1 &35.4 \\
EU & 26.8 &37.9 &\bf53.1 &40.5 &51.2 &37.2 \\
NA & 29.2 &36.4 &39.2 &\bf54.5 &47.1 &39.5 \\
OC & 29.7 &35.1 &30.4 &33.1 &\bf55.9 &31.3 \\
SA & 32.8 &36.2 &36.3 &34.8 &44.8 &\bf51.9 \\
\midrule
\it Performance gap & 22.7 & 17.3 & 25.6 & 22.7 &20.3 &20.6  \\
\bottomrule[1.5pt]
\end{tabular}
}
\end{center}
\label{tab:openearthmap_crossdomain}
\end{table}

\begin{table}[t!]
\color{black}
\caption{
\color{black}
Results of semantic segmentation model~\citep{deeplabv2} trained and evaluated on images from different climate zone on the proposed CASID.
TemMs: temperate monsoon, SubMs: subtropical monsoon, TroMs: tropical monsoon, TroRf: tropical rainforest.
}
\vspace{-2.8mm}
\begin{center}
\scalebox{0.8}{
\begin{tabular}{l|c|c|c|cc}
\toprule[1.5pt]
\diagbox{Train on}{mIoU}{Test on} & TemMs set & SubMs set&TroMs set& TroRf set \\
\midrule
TemMs set & \bf 39.9 &9.4 & 10.9 & 12.2\\ 
SubMs set & 10.0 &\bf 60.9 & 12.8 & 15.3 \\
TroMs set & 12.2 & 12.8 &\bf 56.5 & 9.1 \\
TroRf set & 8.5 & 10.2 & 12.8 &\bf 49.0\\
\midrule
\it Performance gap & 31.4 & 51.5 & 45.6 & 39.9  \\
\bottomrule[1.5pt]
\end{tabular}
}
\end{center}
\label{tab:casid_crossdomain}
\end{table}

\begin{table*}[t!]
\color{black} 
\begin{center}
\vspace{-0mm}
\caption{
\color{black} 
Unsupervised domain adaptation results on the test set of the CASID dataset.
Bkgd: Background, Bldg: Building, Frst: Forest, Rd: Road, Wtr: Water.
All experiments are repeated five times, and the mean (mIoU) and variance ($\pm$) results are presented.
The "*" indicates DAFormer~\citep{2022daformer} using stronger SegFormer~\citep{2021segformer} as the segmentation model.
All other methods, including "target/source only," use the standard DeepLabV2~\citep{deeplabv2} model.
}
\label{tab:DA}
\vspace{-0mm}
\scalebox{0.5918}{
\begin{tabular}{c||c}
\toprule[1.5pt]
\begin{tabular}{c|c|ccccc|cc}
\multirow{2}{*}{Method} & Domain  
& \multicolumn{5}{c|}{IoU (\%)} & \multirow{2}{*}{mIoU (\%)} & \multirow{2}{*}{$\pm$ (\%)} \\
\cline{3-7}
& Source $\rightarrow$ Target
&Bkgd &Bldg &Frst &Rd &Wtr & \\
\midrule
Target only & TemMs $\rightarrow$ TemMs  & 49.4 & 60.9 & 62.6 & 24.7 & 1.9 & 39.9 & 0.4 \\ 
\midrule
Source only & \multirow{8}{*}{SubMs $\rightarrow$ TemMs}
& 0.3 & 8.3 & 41.2 & 0.1 & 0.4 & 10.0 & 0.5 \\ 
AdaptSeg~\citep{adaptseg} & & 44.4 & 60.2 & 45.3 & 21.9 & 6.4 & 35.7 & 1.4 \\ 
CBST~\citep{CBST} & & 37.7 & 54.0 & 23.8 &\bf 27.7 & 4.3 & 29.5 & 1.7 \\ 
CLAN~\citep{CLAN} & & 41.1 & 59.7 & 36.0 & 22.6 & 7.2 & 33.3 & 0.6 \\ 
PyCDA~\citep{PyCDA} & & 37.3 & 32.3 & 22.8 & 7.2 & 0.3 & 20.0 & 0.3 \\ 
FADA~\citep{FADA} & & \bf 48.4 & 60.1 & 59.8 & 19.5 & 5.0 & 38.6 & 0.4 \\ 
ASA~\citep{2020ASA} & & 34.2 & 28.2 &\bf 60.0 & 9.9 & 4.9 & 27.4 & 1.9 \\ 
DAFormer*~\citep{2022daformer} & & 40.1 &\bf 64.4 & 30.0 & 45.1 &\bf 18.2 &\bf 39.6 & 1.1 \\ 
\midrule

Source only & \multirow{8}{*}{TroMs $\rightarrow$ TemMs} 
& 0.5 & 3.1 &\bf 56.6 & 0.0 & 0.6 & 12.2 & 0.2 \\ 
AdaptSeg~\citep{adaptseg} & & 42.3 & 44.1 & 29.7 & 25.1 & 0.7 & 28.4 & 3.1 \\ 
CBST~\citep{CBST} & & 45.2 & 39.1 & 27.2 & 28.0 & 0.3 & 28.0 & 2.0 \\ 
CLAN~\citep{CLAN} & & 41.8 & 50.1 & 34.9 & 26.6 &\bf 1.7 & 31.0 & 2.0 \\ 
PyCDA~\citep{PyCDA} & & 37.7 & 32.0 & 19.6 & 1.9 & 0.2 & 18.3 & 0.2 \\ 
FADA~\citep{FADA} & &\bf 45.7 & 47.1 & 41.3 & 19.3 & 0.6 & 30.8 & 0.3 \\ 
ASA~\citep{2020ASA} & & 40.0 & 31.7 & 36.9 & 1.8 & 0.2 & 22.1 & 1.9 \\ 
DAFormer*~\citep{2022daformer} & & 42.2 &\bf 68.6 & 38.4 &\bf 43.3 & 0.7 &\bf 40.0 & 1.2 \\ 
\midrule

Source only & \multirow{8}{*}{TroRf $\rightarrow$ TemMs} 
 & 3.1 & 3.0 & 35.8 & 0.1 & 0.5 & 8.5 & 0.3 \\ 
AdaptSeg~\citep{adaptseg} & & 38.1 & 59.2 &\bf 65.4 & 24.2 & 4.2 & 38.2 & 0.9 \\ 
CBST~\citep{CBST} & & 36.1 & 57.5 & 59.2 & 27.9 & 1.3 & 36.4 & 2.1 \\ 
CLAN~\citep{CLAN} & & 37.6 & 60.3 & 51.9 & 25.5 & 3.5 & 35.8 & 0.8 \\ 
PyCDA~\citep{PyCDA} & & 30.6 & 20.3 & 25.7 & 1.2 & 0.2 & 15.6 & 0.2 \\ 
FADA~\citep{FADA} & & 40.1 & 57.1 & 64.9 & 19.3 & 1.4 & 36.6 & 0.1 \\ 
ASA~\citep{2020ASA} & & 37.4 & 18.7 & 53.0 &\bf 46.0 & 4.0 & 21.9 & 2.2 \\ 
DAFormer*~\citep{2022daformer} & &\bf 45.2 &\bf 66.1 & 56.5 & 39.3 &\bf 16.8 &\bf 44.8 & 1.0 \\ 

\midrule
\midrule
Target only  & SubMs $\rightarrow$ SubMs & 56.0 & 78.2 & 83.8 & 24.4 & 62.3 & 60.9 & 0.1 \\ 
\midrule

Source only & \multirow{8}{*}{TemMs $\rightarrow$ SubMs}
& 0.7 & 2.8 & 41.4 & 0.1 & 1.9 & 9.4 & 0.9 \\ 
AdaptSeg~\citep{adaptseg} & & 40.3 & 76.6 & 74.8 & 22.5 & 51.7 & 53.2 & 0.9 \\ 
CBST~\citep{CBST} & & 40.9 & 75.7 & 66.3 & 25.8 & 29.7 & 47.7 & 4.5 \\ 
CLAN~\citep{CLAN} & & 45.5 & 76.9 & 77.4 & 23.5 & 39.7 & 52.6 & 0.6 \\ 
PyCDA~\citep{PyCDA} & & 13.3 & 57.5 & 44.0 & 4.4 & 17.9 & 27.4 & 0.8 \\ 
FADA~\citep{FADA} & & 40.0 & 75.6 & 70.9 & 21.9 & 49.1 & 51.5 & 0.5 \\ 
ASA~\citep{2020ASA} & & 24.2 & 53.0 & 62.6 &\bf 66.0 & 13.7 & 30.8 & 3.6 \\ 
DAFormer*~\citep{2022daformer} & &\bf 61.5 &\bf 80.1 &\bf 84.9 & 39.3 &\bf 62.4 &\bf 65.6 & 1.0 \\ 
\midrule

Source only & \multirow{8}{*}{TroMs $\rightarrow$ SubMs} 
& 1.5 & 17.2 & 43.2 & 0.0 & 2.2 & 12.8 & 0.5 \\ 
AdaptSeg~\citep{adaptseg} & & 50.5 & 75.7 & 73.2 & 27.7 & 37.0 & 52.8 & 1.7 \\ 
CBST~\citep{CBST} & & 46.6 & 70.7 & 64.7 & 32.5 & 35.6 & 50.0 & 4.1 \\ 
CLAN~\citep{CLAN} & & 50.2 & 77.9 &\bf 73.8 & 29.8 & 39.1 & 54.1 & 0.8 \\ 
PyCDA~\citep{PyCDA} & & 37.1 & 73.0 & 44.9 & 4.5 & 9.3 & 33.8 & 0.2 \\ 
FADA~\citep{FADA} & & 54.1 & 73.8 & 80.6 & 16.6 & 46.7 & 54.4 & 0.2 \\ 
ASA~\citep{2020ASA} & & 41.2 & 65.6 & 69.4 & 1.1 & 16.7 & 38.8 & 2.1 \\ 
DAFormer*~\citep{2022daformer} & &\bf 54.8 &\bf 81.6 & 73.5 &\bf 44.8 &\bf 49.5 &\bf 60.9 & 0.9 \\ 
\midrule

Source only & \multirow{8}{*}{TroRf $\rightarrow$ SubMs}
& 1.8 & 5.5 & 41.1 & 0.0 & 2.7 & 10.2 & 0.1 \\ 
AdaptSeg~\citep{adaptseg} & & 28.9 & 75.3 & 72.2 & 22.8 &\bf 60.1 & 51.9 & 1.2 \\ 
CBST~\citep{CBST} & & 42.9 & 77.5 & 72.5 & 32.0 & 27.8 & 50.5 & 2.1 \\ 
CLAN~\citep{CLAN} & & 27.0 & 76.8 & 70.8 & 26.6 & 48.2 & 49.9 & 0.9 \\ 
PyCDA~\citep{PyCDA} & & 35.2 & 69.5 & 42.3 & 4.0 & 8.0 & 31.8 & 0.7 \\ 
FADA~\citep{FADA} & & 24.5 & 74.2 & 70.9 & 18.4 & 38.9 & 45.4 & 0.5 \\ 
ASA~\citep{2020ASA} & & 35.5 & 63.7 & 71.4 & 0.0 & 1.6 & 34.4 & 1.9 \\ 
DAFormer*~\citep{2022daformer} & &\bf 50.4 &\bf 80.6 &\bf 77.0 &\bf 45.1 & 59.9 &\bf 62.6 & 1.7 \\ 
\end{tabular}

&
\begin{tabular}{c|ccccc|cc}
Domain 
& \multicolumn{5}{c|}{IoU (\%)} & \multirow{2}{*}{mIoU (\%)} & \multirow{2}{*}{$\pm$ (\%)}\\
\cline{2-6}
Source $\rightarrow$ Target 
&Bkgd &Bldg &Frst &Rd &Wtr & \\
\midrule
TroMs $\rightarrow$ TroMs  & 68.9 & 59.8 & 78.5 & 14.0 & 61.4 & 56.5 & 0.2 \\ 
\midrule
\multirow{8}{*}{TemMs $\rightarrow$ TroMs}
& 1.2 & 1.1 & 48.8 & 0.2 & 3.0 & 10.9 & 0.8 \\ 
& 59.5 & 53.1 & 75.7 & 2.9 & 36.6 & 45.6 & 3.8 \\ 
& 70.1 & 53.2 & 81.0 & 2.3 & 39.5 & 49.2 & 2.6 \\ 
& 64.0 & 56.2 & 81.9 & 3.5 & 42.6 & 49.7 & 0.3 \\ 
& 59.5 & 20.3 & 47.8 & 0.6 & 19.1 & 29.5 & 0.6 \\ 
& 62.5 & 48.8 & 75.4 & 4.5 & 29.8 & 44.2 & 1.1 \\ 
& 50.9 & 19.8 & 70.3 & 0.5 & 51.7 & 38.6 & 1.6 \\ 
&\bf 73.2 &\bf 61.4 &\bf84.0 &\bf 34.2 &\bf 73.0 &\bf 65.2 & 1.4 \\ 
\midrule

\multirow{8}{*}{SubMs $\rightarrow$ TroMs} 
& 1.1 & 4.8 & 53.3 & 0.1 & 4.7 & 12.8 & 0.4 \\ 
& 71.3 & 55.5 & 81.1 & 10.6 & 72.4 & 58.2 & 1.2 \\ 
& 72.1 & 54.0 &\bf 82.0 & 16.5 & 68.7 & 58.7 & 7.7 \\ 
&72.3 & 56.3 & 81.2 & 13.6 & 67.0 & 58.1 & 1.0 \\ 
& 70.8 & 29.1 & 70.2 & 1.2 & 52.7 & 44.8 & 2.0 \\ 
&\bf 72.5 & 50.1 & 80.6 & 7.9 & 46.5 & 51.6 & 0.5 \\ 
& 56.6 & 30.4 & 67.1 & 0.3 & 16.7 & 34.2 & 2.8 \\ 
& 69.2 &\bf 67.8 & 77.8 &\bf 37.2 &\bf 74.6 &\bf 65.3 & 1.5 \\ 
\midrule

\multirow{8}{*}{TroRf $\rightarrow$ TroMs} 
& 2.4 & 1.8 & 43.9 & 0.1 & 6.1 & 10.9 & 0.4 \\ 
& 65.7 & 56.1 &\bf 79.2 & 7.0 & 74.8 & 56.5 & 0.4 \\ 
&\bf 66.3 & 61.2 & 78.9 & 6.8 & 68.9 & 56.4 & 2.1 \\ 
&\bf 66.3 & 56.7 & 78.7 & 5.4 & 76.6 & 56.8 & 0.4 \\ 
& 64.0 & 16.6 & 46.6 & 0.4 & 57.4 & 37.0 & 2.7 \\ 
& 52.0 & 42.9 & 73.6 & 6.5 & 69.2 & 48.8 & 1.0 \\ 
& 52.5 & 28.2 & 66.1 & 5.0 & 33.5 & 36.1 & 2.7 \\ 
& 63.1 &\bf 68.6 & 73.7 &\bf 33.0 &\bf 79.4 &\bf 63.5 & 1.8 \\ 

\midrule
\midrule
TroRf $\rightarrow$ TroRf  & 15.0 & 70.0 & 92.8 & 10.8 & 56.4 & 49.0 & 0.1 \\ 
\midrule
\multirow{8}{*}{TemMs $\rightarrow$ TroRf}
& 0.4 & 0.6 & 58.7 & 0.1 & 1.2 & 12.2 & 2.5 \\ 
& 7.5 &\bf 66.2 & 83.3 & 5.7 & 14.5 & 35.4 & 1.1 \\ 
& 6.5 & 66.0 & 58.1 & 18.7 & 5.6 & 31.0 & 2.0 \\ 
& 8.6 & 64.3 & 81.5 & 8.9 & 28.0 & 38.2 & 0.5 \\ 
& 4.8 & 31.4 & 46.8 & 0.8 & 4.9 & 17.8 & 4.7 \\ 
& 7.1 & 48.0 & 67.4 & 8.1 & 11.4 & 28.4 & 0.7 \\ 
& 7.6 & 29.8 & 83.0 & 1.8 & 0.1 & 24.5 & 2.1 \\ 
&\bf 17.5 & 53.4 &\bf 87.1 &\bf 31.9 &\bf 42.2 &\bf 46.4 & 1.6 \\ 
\midrule

\multirow{8}{*}{SubMs $\rightarrow$ TroRf} 
& 0.4 & 2.6 & 70.1 & 0.1 & 3.5 & 15.3 & 0.7 \\ 
& 26.2 & 70.5 & 91.3 & 11.0 & 21.1 & 44.0 & 0.9 \\ 
& 24.6 & 71.6 & 92.2 & 25.4 & 20.4 & 46.8 & 1.2 \\ 
& 26.1 & 71.6 & 93.2 & 14.7 & 34.8 & 48.1 & 1.7 \\ 
& 3.9 & 17.5 & 45.7 & 1.5 & 9.7 & 15.6 & 0.9 \\ 
& 20.0 & 65.7 & 90.5 & 12.5 & 28.0 & 43.3 & 0.2 \\ 
& 12.5 & 50.1 & 87.0 & 0.5 & 11.9 & 32.4 & 1.2 \\ 
&\bf 31.9 &\bf 75.2 &\bf 94.0 &\bf 40.4 &\bf 75.5 &\bf 63.4 & 1.7 \\ 
\midrule

\multirow{8}{*}{TroMs $\rightarrow$ TroRf} 
 & 0.7 & 0.4 & 41.2 & 0.0 & 2.9 & 9.1 & 0.4 \\ 
& 17.9 & 58.2 & 85.0 & 18.2 & 29.1 & 41.7 & 5.1 \\ 
& 17.1 & 56.9 & 85.1 & 22.1 & 22.0 & 40.6 & 3.5 \\ 
& 18.2 & 69.9 & 85.4 & 20.3 & 29.0 & 44.6 & 0.9 \\ 
& 4.6 & 18.6 & 53.6 & 1.5 & 17.2 & 19.1 & 2.0 \\ 
& 19.0 & 65.2 & 87.4 & 19.9 & 22.3 & 42.8 & 0.1 \\ 
& 9.1 & 50.4 & 81.5 & 2.6 & 23.8 & 33.5 & 2.3 \\ 
&\bf 25.2 &\bf 74.2 &\bf 91.2 &\bf 41.1 &\bf 57.5 &\bf 57.8 & 1.5 \\ 

\end{tabular}\\
\bottomrule[1.5pt]
\end{tabular}
}
\end{center}
\end{table*}

\begin{figure*}[t!]
\begin{center}
\scalebox{1.0}{
\begin{tabular}{cccccc}
\hspace{-3mm}
\includegraphics[width=0.98\linewidth]{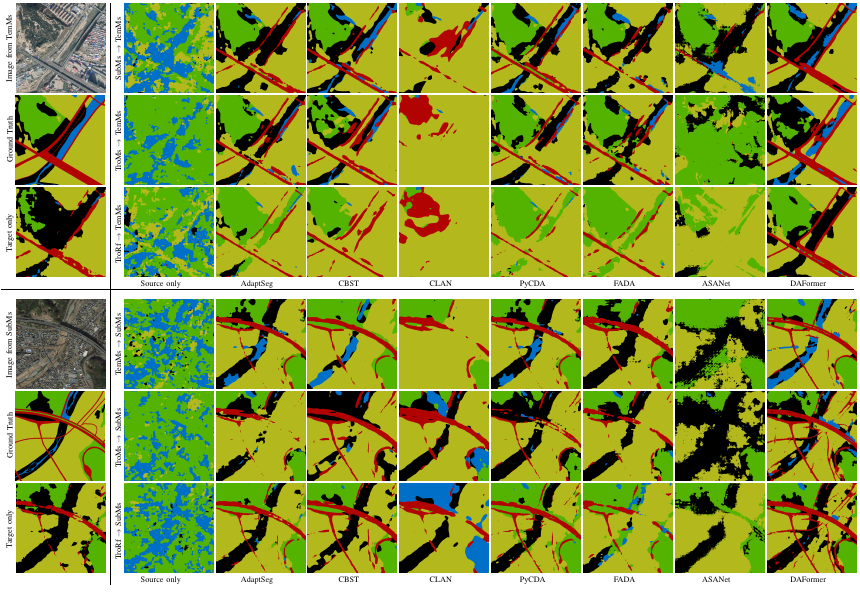}
\end{tabular}
}
\caption{
\color{black}
Visual results for the unsupervised domain adaptation experiments.
The ``A $\rightarrow$ B" indicates that the model is trained on domain A and evaluated on domain B. 
The top three rows are evaluated on the test set of TemMs, and the bottom three rows are evaluated on the test set of SubMs.
}
\vspace{-8mm}
\label{fig:da1}
\end{center}
\end{figure*}

\begin{figure*}[t!]
\begin{center}
\scalebox{1.0}{
\begin{tabular}{cccccc}
\hspace{-3mm}
\includegraphics[width=0.98\linewidth]{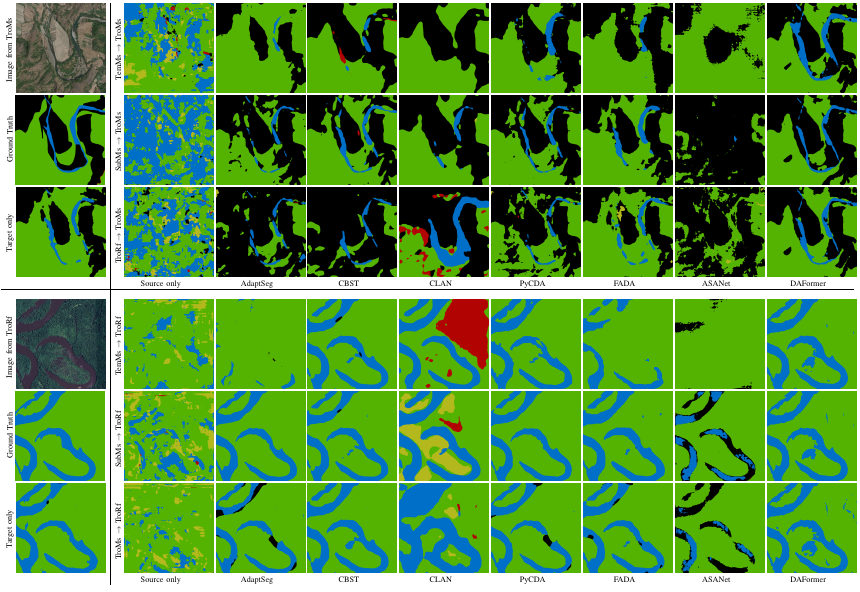}
\end{tabular}
}
\caption{
\color{black}
Visual results for the unsupervised domain adaptation experiments.
The ``A $\rightarrow$ B" indicates that the model is trained on domain A and evaluated on domain B. 
The top three rows are evaluated on the test set of TroMs, and the bottom three rows are evaluated on the test set of TroRf.
}
\vspace{-8mm}
\label{fig:da2}
\end{center}
\end{figure*}

\section{Experiments for Land-Cover Semantic Segmentation}
\label{sec:ss}
{
In this section, we evaluate the following state-of-the-art semantic segmentation approaches to provide baseline results on the newly created CASID dataset. 
We choose some representative methods that are commonly used in remote sensing. 
Specifically, the selected networks are:
}
FCN8S~\citep{fcn2015}, DeepLabV3+~\citep{deeplabv3plus}, PAN~\citep{2018pan}, U-Net~\citep{unet}, U-Net++~\citep{unet++}, PSPNet~\citep{pspnet}, LinkNet~\citep{linknet},  FarSeg~\citep{2020foreground}, FactSet~\citep{factseg}, HRNet~\citep{hrnet}, 
SegFormer~\citep{2021segformer}, PoolFormer~\citep{poolformer}, and SegNeXt~\citep{segnext}.

\subsection{Evaluation Metric}
Following the common practice in~\citep{fcn2015, deeplabv3plus}, we use the intersection over union (IoU) to evaluate the semantic segmentation accuracy. 
{
In terms of IoU for each class, the mIoU represents the mean of the IoUs across all categories.} 
Additionally, we also provide the Dice results.

\subsection{Implementation details}
{
The data split is shown in Table~\ref{tab:cities}. 
During training, we use the Stochastic Gradient Descent (SGD) optimizer with a momentum of 0.9 and a weight decay of 1e-4. 
The learning rate was initially set to 1e-2, and a poly schedule~\citep{deeplabv3plus} with a power of 0.9 is applied. 
For recent models like SegFormer~\citep{2021segformer}, PoolFormer~\citep{poolformer}, and SegNeXt~\citep{segnext}, we adopt their original training settings. 
For example, SegFormer is trained using AdamW~\citep{2017adamw} with an initial learning rate of 6e-5. 
The number of training iterations for all models is set to 40k with a batch size of 8. 
Data augmentation is applied by randomly cropping $512\times 512$ patches from the original images, along with random mirroring and rotation. 
Following~\citep{fcn2015, deeplabv3plus, 2021loveda}, the backbones used in all networks are pre-trained on ImageNet~\citep{imagenet2015}. 
We conduct all experiments five times and present the mean and variance results.
}

{\color{black}
\subsection{Quantitative Results}
The quantitative results are provided in { Table~\ref{tab:ss}}.
We can see the categories of road and water are more challenging than building and forest, due to their thin and long shapes.
The FCN8S~\citep{fcn2015} and PAN~\citep{2018pan} with a shallow backbone of VGG-16~\citep{vgg} and ResNet-50~\citep{resnet2016} show relatively low 
performance, \ie, 57.2 mIoU and 56.0 mIoU.
By using multi-scale context information, deeplabV3+~\citep{deeplabv3plus} and PSPNet~\citep{pspnet} achieve 56.9 mIoU and 57.1 mIoU.
FarSeg~\citep{2020foreground} and FactSeg~\citep{factseg} improve the semantic segmentation model to pay attention to the foreground of remote sensing images, and their performance reaches 58.2 mIoU and 58.0 mIoU.
With a stronger backbone of HRNet-W32~\citep{hrnet}, HRNet~\citep{hrnet} achieves better performance, \ie, 60.9 mIoU, especially for road categories.
U-Net~\citep{unet}, U-Net++, and LinkNet~\citep{linknet} enhance the semantic segmentation model by utilizing detail information in shallow features. 
Thus they achieve much better accuracy for categories of road and water, which require detail information on the most.
Among them, the LinkNet~\citep{linknet} achieves a high accuracy of 62.3 mIoU.
By utilizing a self-attention mechanism, recent transformer-based architectures such as SegFormer~\citep{2021segformer} and PoolFormer~\citep{poolformer} have shown competitive results in land-cover semantic segmentation, achieving 63.0 and 62.7 mIoU, respectively. 
In contrast, the recent CNN-based SegNeXt~\citep{segnext} employs efficient convolutional attention and has achieved the highest results of 63.4 mIoU.
}

\subsection{Qualitative Results}
{
Some representative results are shown in Figures~\ref{fig:seg1}, \ref{fig:seg2}, and~\ref{fig:seg3}. 
The road category proves to be the most challenging, as most methods fail to accurately predict it. 
In contrast, the forest category achieves higher accuracies across all methods. 
This can be attributed to the forest's strong spectral homogeneity (low intra-class variance) and distinct spectral features compared to other categories (large inter-class variance), as shown in Figure~\ref{fig:dataset_rgb}. 
Moreover, the road's thin and elongated shape makes it susceptible to being obscured by roadside trees or shadows from buildings.

With the shallow backbone VGG-16~\citep{vgg}, FCN8S~\citep{fcn2015} struggles to recognize the road and water due to its limited feature extraction capability. 
However, other methods utilizing more advanced backbones~\citep{resnet2016, hrnet} and segmentation architectures~\citep{deeplabv3plus, pspnet, unet, unet++, linknet, hrnet, factseg, 2020foreground} produce improved results. 
The qualitative results align with the quantitative findings, with HRNet~\citep{hrnet} demonstrating the best visualization of the road due to its high-resolution retention and multi-scale fusion.
The recent transformer-based architectures, SegFormer~\citep{2021segformer} and PoolFormer~\citep{poolformer}, along with the CNN-based SegNeXt~\citep{segnext}, deliver competitive results, as depicted in Figure~\ref{fig:seg3}.
}



\section{Experiments for Unsupervised Domain Adaptation}
\label{sec:uda}
In this section, we evaluate the advanced Unsupervised Domain Adaptation (UDA) methods on the CASID dataset.
We choose two mainstream UDA approaches, \ie, adversarial training (AdaptSeg~\citep{adaptseg}, CLAN~\citep{CLAN}, FADA~\citep{FADA}) and self-training (CBST~\citep{CBST}, PyCDA~\citep{PyCDA}). 
{\color{black}
Recent methods ASA~\citep{2020ASA} and DAFormer~\citep{2022daformer} are also included.
The evaluation uses the same metrics as that of the semantic segmentation task~\citep{fcn2015, deeplabv3plus}, \ie, mean intersection over union (mIoU).
We conduct all experiments five times and present the mean and variance results.
} 

\subsection{Implementation details}
All UDA methods adopt the same feature extractor and discriminator, following the common practice in~\citep{FADA, adaptseg}. 
{
By default, if not specified, DeepLabV2~\citep{deeplabv2} with ResNet-50~\citep{resnet2016} is used as the feature extractor, and the discriminator is constructed using fully convolutional layers.

For the adversarial training methods, the learning rates for classification and discriminator are set to 5e-3 and 1e-4, respectively. 
The Adam optimizer is employed for the discriminator with momentum values of 0.9 and 0.99. The number of training iterations is set to 10k, with a batch size of 8. Source images and target images are alternately input to the model.
The learning schedule and other settings are the same as in semantic segmentation settings. 
For the self-training methods, the learning rate for classification is set to 1e-2. 
For recent methods like ASA~\citep{2020ASA} and DAFormer~\citep{2022daformer}, we adopt their original training settings. 
For example, for DAFormer~\citep{2022daformer}, we utilize AdamW as the optimizer, with a learning rate of 6e-5, and 6e-4 for the encoder, and decoder.
}

\subsection{Quantitative Results}
\vspace{+1mm}
\noindent{\bf{Dataset comparison. }}
We provide results reported on the two most relevant existing datasets, \ie, LoveDA~\citep{2021loveda} and OpenEarthMap~\citep{2023openearthmap}, and compare their cross-domain performance gap with the proposed CASID datasets.
The cross-domain performance gap is the performance difference between cross-domain results (train on a source domain A and test on a different target domain B) and target-only results (train and test on the same domain).
It demonstrates the performance degradation caused by the domain gap in the datasets.
A higher cross-domain performance gap indicates a larger domain gap between different domains in the dataset, which makes the dataset more challenging.
As shown in Tables~\ref{tab:loveda_crossdomain},~\ref{tab:openearthmap_crossdomain}, and~\ref{tab:casid_crossdomain}, cross-domain performance gap in the LoveDA~\citep{2021loveda} and OpenEarthMap~\citep{2023openearthmap} are 14.4$\sim$15.1 mIoU and 17.3$\sim$25.6 mIoU, respectively.
{
In contrast, the cross-domain performance gap in the proposed CASID dataset is 31.4$\sim$51.5, which is much higher than existing datasets. 
}
This demonstrates that the proposed CASID dataset can serve as a more challenging benchmark for both semantic segmentation and multi-domain unsupervised domain adaptation tasks.

{\color{black}
Moreover, in the existing datasets LoveDA~\citep{2021loveda} and OpenEarthMap~\citep{2023openearthmap}, samples are simply divided into different domains based on geographical location.
On the contrary, the proposed CASID is climate-aware and collected from 30 areas in different climate zone, which is unique.
Notice that CASID can also be divided based on geographical location if needed.
The challenging and distinct characteristics of the CASID dataset can be highly beneficial to the computer vision and remote sensing community, and accelerate progress in relevant research.
} 

\vspace{+1mm}
\noindent{\bf{UDA results. }}
{As shown in Table~\ref{tab:DA}}, when the target domain is TemMs set, the results of source only (without UDA methods) are extremely low compared with the result of target only, \eg, the results of SubMs $\rightarrow$ TemMs, TroMs $\rightarrow$ TemMs, and TroMs $\rightarrow$ TemMs are 10.0 mIoU, 12.2 mIoU, and 8.5 mIoU, respectively.
While the result of target only is 39.9 mIoU.

When we evaluate on SubMs set, TroMs set, and TroRf set, results are similar, \ie, land-cover semantic segmentation models trained on a certain climate zone significantly lose their effectiveness on new images captured from different climate zones.
This is caused by the climate domain gap as we analyze in Section~\ref{sec:climate_domain_gap}.
After using UDA methods~\citep{adaptseg, CBST, CLAN, PyCDA, FADA, 2020ASA, 2022daformer}, the cross-domain results are largely improved, \eg, SubMs $\rightarrow$ TemMs with AdaptSeg~\citep{adaptseg} achieves 35.7 mIoU, { which is comparable to the result of TemMs $\rightarrow$ TemMs.}

{
Meanwhile, when using DeepLabV2~\citep{deeplabv2}, the best result of TroMs $\rightarrow$ TemMs is 31.0 mIoU, which is obtained by CALN~\citep{CLAN}. 
However, this is still far from the result of TemMs $\rightarrow$ TemMs, which achieves 39.9 mIoU. This indicates that the domain gap of TroMs $\rightarrow$ TemMs is larger than that of SubMs $\rightarrow$ TemMs.
Although DAFormer~\citep{2022daformer} achieves a slightly higher mIoU of 40.0 compared to the TemMs $\rightarrow$ TemMs target only with DeepLabV2~\citep{deeplabv2}, it is important to note that it uses the strong SegFormer-B5~\citep{2021segformer} as the segmentation model, which is not a fair comparison.
This indicates that the climate domain gap leads to a higher computation cost to maintain accuracy.
}

When we evaluate on the SubMs set, TroMs set, and TroRf set, the results are also similar, meaning that UDA methods~\citep{adaptseg, CBST, CLAN, PyCDA, FADA, 2020ASA} can significantly improve the results.
{
However, in cases where the domain gap is relatively large, the results after UDA are still far from the results of target only. 
For example, TroRf $\rightarrow$ SubMs with FADA~\citep{adaptseg} only achieves 54.4 mIoU, while SubMs $\rightarrow$ SubMs with target only achieves 60.9 mIoU.
This indicates that existing UDA methods still have room for improvement.
}

\vspace{+1mm}
\noindent{\bf{Intra-domain variance (IDV). }}
{
Additionally, we conduct further experiments to analyze the intra-domain variance in our dataset. 
We select TemMs/TroRf as the source/target domain because they exhibit larger differences in climate and domain characteristics.

We notice that intra-domain variance mainly arises from the number of sampling regions. 
In general, samples from the same regions are more similar. Their distance after t-SNE is small, and they usually belong to the same cluster, while samples from different regions exhibit greater dissimilarity and belong to different clusters, thus substantially increasing the variance.
Therefore, having fewer sampling areas results in lower IDV. 
To ensure that reducing the source samples leads to a decrease in IDV, we follow the sample reduction protocol shown in Table~\ref{tab:reducing_cities}. 
Instead of uniformly reducing the source samples, we reduce sample regions.
We measure the IDV by calculating the standard deviation, $\sigma$, in t-SNE results.
We denote these versions as TemMs-100\%, TemMs-50\%, and TemMs-25\% to indicate the reduction percentages.
We observe a gradual reduction in variance $\sigma$ as we decrease the sample regions (from $\sigma=0.64$ for TemMs-100\% to $\sigma=0.32$ for TemMs-25\%), as shown in Table~\ref{tab:intra-domain-variance}.
This indicates that we reduce the IDV by reducing the sample regions.
As a result, the domain adaptation performance of all recent state-of-the-art methods dramatically degrades from TemMs-100\% to TemMs-25\%.
This demonstrates that a higher intra-domain variance benefits the UDA task.

To eliminate any interference caused by differences in sample quantity, we have conducted a control experiment where we uniformly select 25\% and 50\% samples in TemMs. We refer to these uniformly sampled versions as TemMs-25\%-U and TemMs-50\%-U, respectively. 
The protocol for TemMs-25/50\%-U is presented in Table~\ref{tab:reducing_samples_uniform}. 
It is important to note that TemMs-25/50\%-U has exactly the same number of samples as TemMs-25/50\%. 
As shown in Table~\ref{tab:intra-domain-variance}, owing to uniform sampling, TemMs-25/50\%-U demonstrate similar variance ($\sigma$) to the full TemMs-100\%, which higher than TemMs-25/50\%.
Accordingly, TemMs-25/50\%-U achieves better performance than TemMs-25/50\%.
Noting that TemMs-25/50\%-U have the same number of samples for training with TemMs-25/50\%, the only main difference between TemMs-25/50\%-U and TemMs-25/50\% is variance.
Therefore, we can conclude that the reduction in UDA performance from TemMs-25/50\%-U to TemMs-25/50\% is not due to a lack of training data from the source domain. Instead, it is attributed to the lower intra-domain variance.
Interestingly, although TemMs-50\%-U has less samples for training, with similar variance, it achieves comparable results to TemMs-100\%.
}

\begin{table}[t!]
\begin{center}
\vspace{-0mm}
\caption{
The protocol for reducing source domain samples to reduce variance.
}
\label{tab:reducing_cities}
\vspace{-0mm}
\scalebox{0.518}{
\begin{tabular}{c|c|c|cc}
\toprule[1.5pt]
Climate & Area & Images & Train/Test split \\
\midrule
\multirow{4}{*}{100\% Temperate monsoon}
& Hokkaido, Japan & 21 
& \multirow{5}{*}{Train: 201} \\

\multirow{4}{*}{(TemMs-100\%)}
& Seoul, South Korea& 55 & \\
& Zhangjiakou, Hebei, China& 35 & \\
& Weifang, Shandong, China& 48& \\
& Daqing, Heilongjiang, China& 42& \\

\midrule 

\multirow{2}{*}{50\% Temperate monsoon}
& Hokkaido, Japan & 21 
& \multirow{3}{*}{Train: 98} \\
\multirow{2}{*}{(TemMs-50\%)}
& Zhangjiakou, Hebei, China& 35 & \\
& Daqing, Heilongjiang, China& 42& \\

\midrule 

\multirow{1}{*}{25\% Temperate monsoon}
& Hokkaido, Japan & 21 
& \multirow{2}{*}{Train: 56} \\
\multirow{1}{*}{(TemMs-25\%)}
& Zhangjiakou, Hebei, China& 35 & \\
\bottomrule[1.5pt]
\end{tabular}}
\end{center}
\end{table}

\begin{table}[t!]
\begin{center}
\vspace{-0mm}
\caption{
The protocol for uniformly reducing source domain samples.
}
\label{tab:reducing_samples_uniform}
\vspace{-0mm}
\scalebox{0.518}{
\begin{tabular}{c|c|c|cc}
\toprule[1.5pt]
Climate & Area & Images & Train/Test split \\
\midrule

\multirow{4}{*}{50\% Temperate monsoon}
& Hokkaido, Japan & 10 
& \multirow{5}{*}{Train: 98} \\

\multirow{4}{*}{(TemMs-50\%-U)}
& Seoul, South Korea& 28 & \\
& Zhangjiakou, Hebei, China& 18 & \\
& Weifang, Shandong, China& 22& \\
& Daqing, Heilongjiang, China& 20& \\

\midrule

\multirow{4}{*}{25\% Temperate monsoon}
& Hokkaido, Japan & 6 
& \multirow{5}{*}{Train: 56} \\

\multirow{4}{*}{(TemMs-50\%-U)}
& Seoul, South Korea& 18 & \\
& Zhangjiakou, Hebei, China& 10 & \\
& Weifang, Shandong, China& 12& \\
& Daqing, Heilongjiang, China& 10& \\
\bottomrule[1.5pt]
\end{tabular}}
\end{center}
\end{table}

\begin{table}[t!]
\vspace{-0mm}
\caption{
Experiments for intra-domain variance (IDV).
The source domain is TemMs, and the target domain is TroRf.
TemMs-25/50\% is created by reducing sampling areas, resulting in lower diversity and lower IDV compared to TemMs-25/50\%-U which is uniformly sampled on the whole TemMs.
A higher standard variance $\sigma$ indicates a higher IDV.
As a result, methods trained on TemMs-25/50\% exhibit lower performance than TemMs-25/50\%-U.
This indicates that higher IDV is beneficial for increasing performance.
}
\begin{center}
\scalebox{0.4918}{
\begin{tabular}{c|cccc|cccc|ccc}
\toprule[1.5pt]
\multirow{3}{*}{Method} & \multicolumn{2}{c|}{TemMs-25\%} & \multicolumn{2}{c|}{TemMs-25\%-U} & \multicolumn{2}{c|}{TemMs-50\%} & \multicolumn{2}{c|}{TemMs-50\%-U} & \multicolumn{2}{c}{TemMs-100\%} \\
& \multicolumn{2}{c|}{$\sigma = 0.32$} & \multicolumn{2}{c|}{$\sigma = 0.60$} & \multicolumn{2}{c|}{$\sigma = 0.45$} & \multicolumn{2}{c|}{$\sigma = 0.62 $} & \multicolumn{2}{c}{$\sigma = 0.64 $}  \\
\cline{2-11}
& mIoU & Dice & mIoU & Dice & mIoU & Dice & mIoU & Dice & mIoU & Dice \\ 

\midrule
Target only (TroRf $\rightarrow$ TroRf) & - & - & - & - & -  & - & - & -  & 49.0 &59.3 \\
\midrule

Source only (TemMs $\rightarrow$ TroRf)
& 7.7 &11.6 &\bf 12.8 &\bf 16.6 & 12.0 & 15.2 &\bf 12.7 &\bf 16.9 & 12.2 & 15.7\\
AdaptSeg~\citep{adaptseg} &30.0 &39.1 &\bf 30.7 &\bf 39.6 &32.3 &41.5 &\bf 32.9 &\bf 41.6 & 35.4 & 59.0 \\
CBST~\citep{CBST} & 28.1 &39.6 &\bf 32.4 &\bf 42.3 & 30.6 & 40.7 &\bf 32.7 &\bf 44.2 & 31.0 & 41.4\\
CLAN~\citep{CLAN} & 31.1& 40.2 &\bf 35.3 &\bf 42.8 &31.2 &40.7 &\bf 36.4 &\bf 46.1 & 38.2 & 48.8\\
PyCDA~\citep{PyCDA} & 11.1 & 18.4 &\bf 16.7 &\bf 25.3 &17.2 & 27.1 &\bf 17.5 &\bf 27.7 & 17.8 & 26.2 \\
FADA~\citep{FADA} & 22.7 &32.5 &\bf  25.8 &\bf 37.1 &23.6 &32.8 &\bf 29.8 &\bf 41.2 & 28.4 & 38.8\\
\midrule
Average & 21.8 & 30.2 &\bf 25.6 &\bf 34.0 & 24.5 & 33.0 &\bf 27.0 &\bf 36.3 & 27.1 & 38.3\\
\bottomrule[1.5pt]
\end{tabular}}
\end{center}
\label{tab:intra-domain-variance}
\end{table}

\begin{figure}[h!]
\color{black}
\centering
\scalebox{0.98}{
\begin{tabular}{cccccccc}
\includegraphics[width=0.918\linewidth]{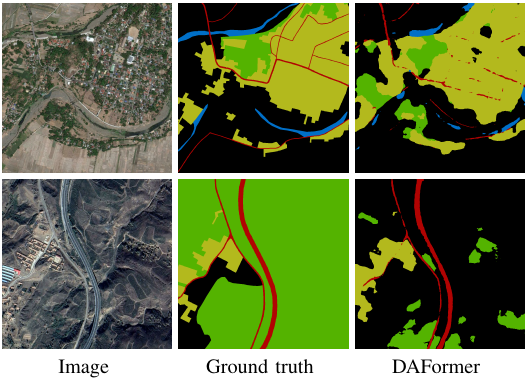}
\end{tabular}}
\caption{
\color{black}
Failure UDA examples on the CASID.
}
\label{fig:failure_uda}
\end{figure}
\begin{figure}[t!]
\color{black}
\begin{center}
\scalebox{1.0}{
\begin{tabular}{cccc}
\includegraphics[width=0.95\linewidth]{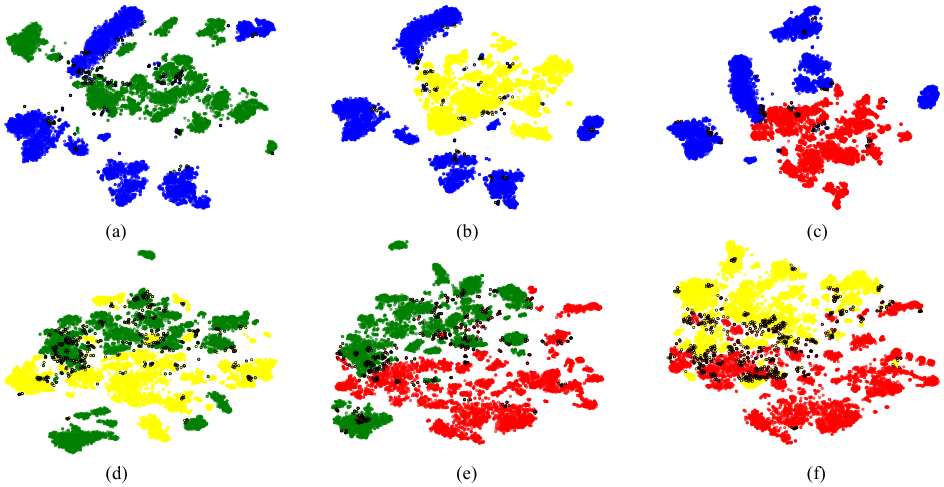} \\
\end{tabular}
}
\caption{
Visualization of the data representations for the source and the target domains.
We presented pairwise two-dimensional t-SNE~\citep{2008tsne} visualizations of 4 domains in the CASID.
Different colors indicate different domains, Blue: TemMs, Green: SubMs, Yellow: TroMs, Red: TroRf.
The symbol ``$\circ$" indicates the marginal data, which are predicted by the k-Nearest Neighbors (k-NN) algorithm.
If a data point's nearest neighbor belongs to a different category, then that data point can be considered one of the marginal points, \ie, they are at the boundary of different domains.
(a) TemMs \& SubMs.
(b) TemMs \& TroMs.
(c) TemMs \& TroRf.
(d) SubMs \& TroMs.
(e) SubMs \& TroRf.
(f) TroMs \& TroRf.
Zoom for better details.
}
\label{fig:source_target_domain_tSNE}
\end{center}
\end{figure}

\subsection{Qualitative Results}

Some representative results are shown in Figures~\ref{fig:da1} and~\ref{fig:da2}.
{
The results of the target-only models successfully recognize most of the buildings and roads, and are the closest to the ground truth. 
On the other hand, the source-only models catastrophically fail to predict accurate results.
This demonstrates that the climate domain gap severely degrades the models trained on a certain climate zone. 
}
After applying UDA methods~\citep{adaptseg, CBST, CLAN, PyCDA, FADA, 2020ASA, 2022daformer}, the qualitative results are greatly improved, which is consistent with Table~\ref{tab:DA}. 
{
However, there is still significant room for improvement. 
For example, the consistency of segmentation is not desirable, and the accuracy for road and water segments is not satisfactory.
More examples of failure cases are shown in Figure~\ref{fig:failure_uda}. It can be observed that large areas of false segmentation still exist, even with the recent strong DAFormer~\citep{2022daformer}.
}
We also provide source-target pairwise t-SNE visualized results in Figure~\ref{fig:source_target_domain_tSNE}.

\section{Conclusion}
\label{sec:conclusion}
In this paper, we investigate the impact of climate on land-cover semantic segmentation for the first time. 
To this end, we introduce a unique large-scale Climate-Aware Satellite Images Dataset (CASID) for domain adaptive land-cover semantic segmentation. 
{
It consists of 980 satellite images with a size of $5,000\times 5,000$ pixels, collected from 30 different regions in four typical climate zones: temperate monsoon, subtropical monsoon, tropical monsoon, and tropical rainforest. 
This makes our CASID the first land-cover semantic segmentation dataset with multiple climate domains. 
}
It not only benefits the research community in advancing domain adaptive land-cover semantic segmentation with higher generalization ability but also shows the potential to become a challenging benchmark for the design of robust vision algorithms beyond the field of remote sensing.
{
Moreover, we conduct a series of statistical analyses on images collected from the four typical climate zones. 
These analyses reveal significant differences in global image features, image texture, category distribution, spectral values, and object shapes among these zones, resulting in a climate domain gap. 
These findings provide valuable insights to inspire subsequent research in this field.
Additionally, we evaluate various state-of-the-art semantic segmentation and unsupervised domain adaptation methods using our CASID dataset. 
These evaluations establish a solid baseline for future research. 
On the basis of these experimental results, we conclude that CASID can serve as a benchmark dataset for evaluating the performance of land-cover semantic segmentation and unsupervised domain adaptation methods.
We hope that this work can benefit the research community by advancing land-cover segmentation with improved generalization ability and inspiring further research in this field.
}

{
In the future, our aim is to explore and develop an unsupervised domain adaptation method to address the domain gap caused by climate, using the CASID dataset as a foundation. 
Moreover, we plan to expand the CASID dataset to include more types of climate zones and collect image samples from a global perspective.
}










\bibliographystyle{unsrtnat}
\bibliography{mybibfile.bib}

\end{document}